\documentclass[10pt,twocolumn,letterpaper]{article}

\usepackage{cvpr}              % To produce the CAMERA-READY version
\usepackage[accsupp]{axessibility}  % Improves PDF readability for those with disabilities.
\usepackage{multirow}
\usepackage{arydshln}
\usepackage{tcolorbox}
\usepackage{colortbl}
\usepackage{xcolor}
\definecolor{darkgreen}{HTML}{008000}
\usepackage{cuted}
\usepackage{footmisc}
\definecolor{cvprblue}{rgb}{0.21,0.49,0.74}
\usepackage[pagebackref,breaklinks,colorlinks,allcolors=cvprblue]{hyperref}

\def\confName{CVPR}
\def\confYear{2026}

\title{MRD: Multi-resolution Retrieval-Detection Fusion for High-Resolution Image Understanding}
\author{
    Fan Yang$^1$, \ Xingping Dong$^2$, \ Xin Yu$^3$, \ Wenhan Luo$^{4}$, \ Wei Liu$^{5}$, \ Kaihao Zhang$ \thanks{Corresponding author.}$ \\
    $^1$HITSZ \quad $^2$Wuhan University \quad $^3$The University of Queensland \quad $^4$ HKUST \quad  $^5$ Video Rebirth
}

\begin{document}
\maketitle
% \footnotetext{$^\dag$ Corresponding author.}

\begin{abstract}
% Understanding high-resolution (HR) images remains a critical challenge for multimodal large language models (MLLM). Recent approaches leverage vision-based retrieval-augmented generation (RAG) models to retrieve query-relevant crops from HR images enhancing the understanding accuracy of MLLMs . However, this paradigm often leads to object fragmentation—inducing semantic bias and incomplete retrieval—and generates false positives from irrelevant background patches, thereby degrading HR image understanding performance. To tackle these limitations, we propose \textbf{Multi-resolution Retrieval-Detection (MRD)}, a training-free framework designed to enhance HR image understanding from both local and global perspectives. Specifically, locally, MRD enforces cross-scale semantic consistency via multi-resolution semantic fusion to rectify single-resolution bias and alleviate object fragmentation; globally, it incorporates open-vocabulary object detection (OVD) as localization priors into a unified scoring framework. Extensive experiments on HR image benchmarks across multiple MLLMs demonstrate that our MRD approach achieves state-of-the-art (SOTA) performance on both single-object and multi-object understanding tasks. Code will be available at:
% \url{https://github.com/username/MRD}
Understanding high-resolution (HR) images remains a critical challenge for multimodal large language models (MLLMs). Recent approaches leverage vision-based retrieval-augmented generation (RAG) to retrieve query-relevant crops from HR images, improving understanding capacity
of MLLMs. However, this paradigm often leads to object fragmentation, resulting in semantic bias and incomplete retrieval, while also introducing false positives from irrelevant background patches. To address these issues, we propose \textbf{Multi-resolution Retrieval-Detection (MRD)}, a training-free framework that enhances HR image understanding from both local and global perspectives. Locally, MRD enforces cross-scale semantic consistency via multi-resolution semantic fusion to mitigate single-resolution bias and alleviate object fragmentation. Globally, it integrates open-vocabulary object detection (OVD) as localization priors within a unified framework. Extensive experiments across multiple MLLMs on HR image benchmarks demonstrate that MRD achieves state-of-the-art (SOTA) performance on both single-object and multi-object understanding tasks. Code will be available at:
\url{https://github.com/yf0412/MRD}.
\end{abstract}

% Understanding high-resolution (HR) images remains a critical challenge for multimodal large language models (MLLMs). Recent approaches leverage vision-based retrieval-augmented generation (RAG) to retrieve query-relevant crops from HR images, improving understanding capacity of MLLMs. However, this paradigm often leads to object fragmentation, resulting in semantic bias and incomplete retrieval, while also introducing false positives from irrelevant background patches. To address these issues, we propose \Multi-resolution Retrieval-Detection (MRD), a training-free framework that enhances HR image understanding from both local and global perspectives. Locally, MRD enforces cross-scale semantic consistency via multi-resolution semantic fusion to mitigate single-resolution bias and alleviate object fragmentation. Globally, it integrates open-vocabulary object detection (OVD) as localization priors within a unified framework. Extensive experiments across multiple MLLMs on HR image benchmarks demonstrate that MRD achieves state-of-the-art (SOTA) performance on both single-object and multi-object understanding tasks. Code will be available at: https://github.com/yf0412/MRD.
    
% \vspace{-4mm}
\section{Introduction}
\label{sec:intro}
\begin{figure}[t]
  \centering
   \includegraphics[width=1\linewidth]{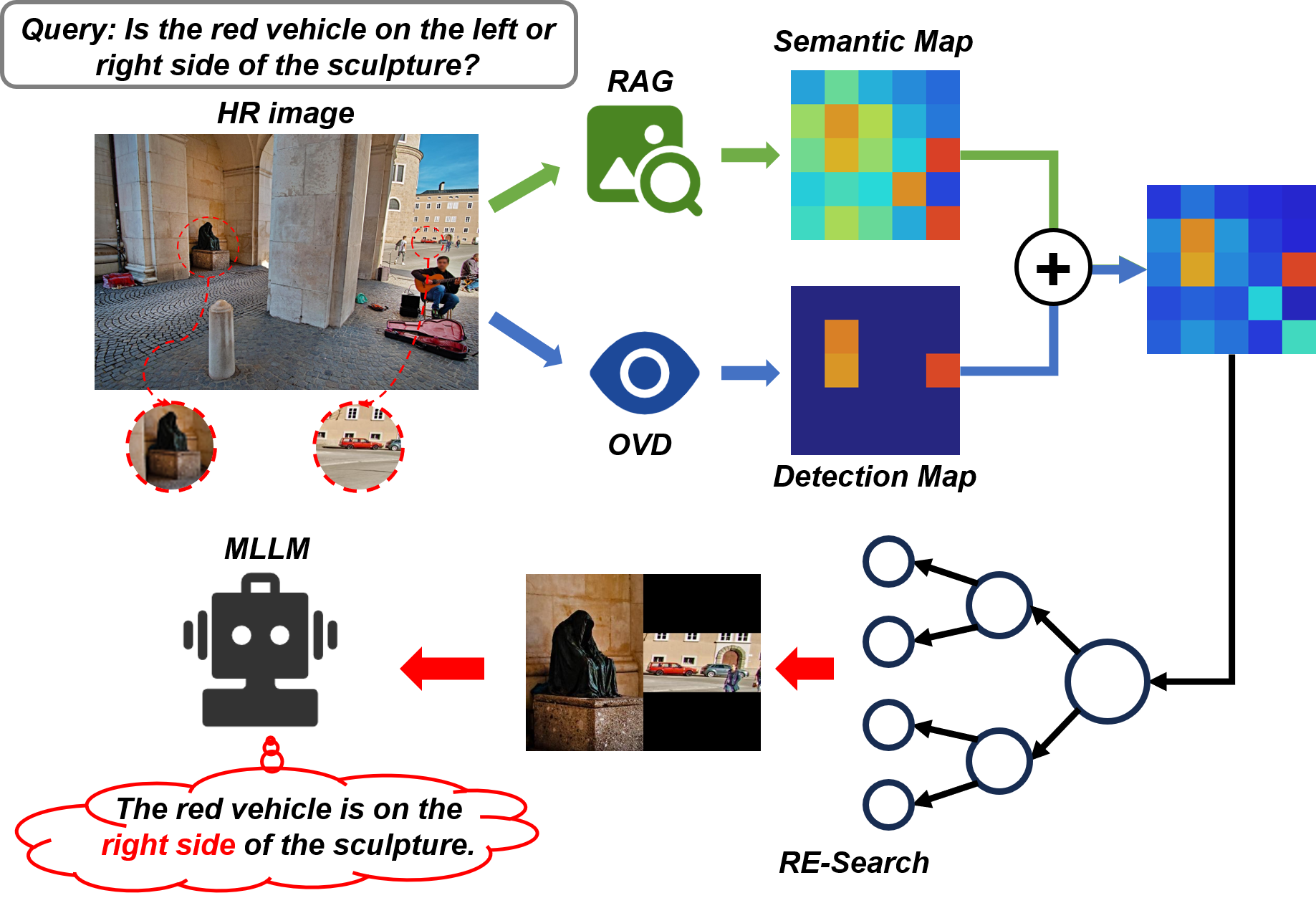}

   \caption{An illustration of the \textbf{\textit{Multi-resolution Retrieval-Detection (MRD)}} framework for HR image understanding.
   % , which uses RAG and OVD to obtain semantic similarity map and detection confidence map respectively. By integrating the two, the target objects can be localized more accurately.
   }
   \label{fig:mrd_short}
\end{figure}
Multimodal large language models (MLLMs) have driven a transformative paradigm shift in cross-modal understanding, achieving remarkable proficiency in aligning visual signals with linguistic semantics and enabling high-level cross-modal reasoning~\cite{yin2024survey, bai2025qwen3, team2024gemini}. By bridging the gap between pixel-level visual data and abstract semantic comprehension, MLLMs have become the foundational architecture for mainstream vision tasks, from visual question answering to complex open-world scene interpretation~\cite{hurst2024gpt, fu2024mme, team2025evaluating}.

However, a critical limitation remains in standard MLLM pipelines: the inadequate capability to effectively handle high-resolution images. Real-world visual data is inherently rich in fine-grained details, and the ability to perceive such features—including small objects, delicate textures, and intricate text—is essential for precise visual comprehension and complex cross-modal reasoning~\cite{wu2024v, zhang2024mme, wang2025divide}.
Most MLLMs exhibit inherent limitations in this regard, hampered by the constraint of fixed low input resolutions~\cite{bai2023qwen,liu2024improved,liu2024llavanext}. Recent studies indicate that existing methods remain unsatisfactory for high-resolution image tasks which is clearly demonstrated by their suboptimal performance on HR benchmarks~\cite{wu2024v,zheng2025deepeyes,zhang2025mllms,shen2025zoomeye}.

% However, a pervasive bottleneck persists in standard MLLM pipelines: the constraint of fixed, low input resolutions (typically $448 \times 448$)~\cite{bai2023qwen,liu2024improved,liu2024llavanext}. While this design choice significantly reduces computational overhead and simplifies architecture, it inevitably introduces aliasing and severe loss of high-frequency details. Consequently, when applying these models to High-Resolution (HR) real-world scenarios, the resizing process often leads to shape distortion and the obliteration of fine-grained visual cues. Recent empirical studies ~\cite{li2025dyfo,wang2025retrieval,shen2025zoomeye} corroborate that state-of-the-art MLLMs exhibit suboptimal performance on HR understanding benchmarks ~\cite{wang2025divide}, struggling to recognize small objects or detailed text.

To address this limitation, enhancing HR image perception of MLLMs has emerged as a prominent research focus. The dominant "localize-and-zoom-in" paradigm improves detail perception by identifying and cropping informative sub-regions in HR images for refined processing, with existing methods falling into two categories. Training-based approaches (e.g., supervised fine-tuning (SFT)~\cite{shao2024visual,wu2024v} or reinforcement learning (RL)~\cite{zhang2025thyme, zheng2025deepeyes}) learn effective region localization, but suffer from prohibitive computational costs, long training cycles, and poor cross-architecture transferability, limiting practical scalability. In contrast, training-free methods~\cite{zhang2025mllms, li2025dyfo, shen2025zoomeye, wang2025retrieval} localize regions via attention maps or heuristic search without parameter updates. However, these top-down methods often fail in initial coarse-grained search: downsampled views lack fine-grained details, leading to irreversible error propagation, and misdirected search. Besides, they also tend to focus on main targets while neglecting others in multi-object tasks.

% To mitigate this limitation, enhancing the high-resolution image perception capability of MLLMs has emerged as a prominent research focus. A widely used "localize-and-zoom" strategy is commonly employed to boost detail perception in these models.
% This strategy involves identifying informative sub-regions within a high-resolution image and cropping them for detailed processing. Existing approaches generally fall into two categories. Training-based methods, such as those employing Supervised Fine-Tuning (SFT) ~\cite{qin2025chain,shao2024visual,wu2024v} or Reinforcement Learning (RL) ~\cite{zhang2025thyme, zheng2025deepeyes, li2025look, lai2025mini, wang2025pixel}, can learn to localize relevant regions effectively. However, they are hampered by prohibitive computational costs, lengthy training cycles, and limited transferability across different model architectures. In contrast, training-free methods ~\cite{li2025dyfo, shen2025zoomeye, wang2025retrieval} attempt to localize regions using attention maps or heuristic search without parameter updates. Despite their flexibility, these top-down approaches often fail in the initial "coarse" search stage. Due to the lack of detail in downsampled views, they frequently miss small objects, leading to irreversible error propagation and incorrect search paths. Furthermore, these methods tend to focus on one object while neglecting other target objects in multi-object tasks.

Inspired by the success of retrieval-augmented generation in long-context large language models (LLMs)~\cite{jin2024long}, recent work Retrieval-Augmented Perception (RAP)~\cite{wang2025retrieval} extends this method to HR image understanding. Specifically, RAP adopts an adaptive tree-search strategy and leverages a vision-based RAG model~\cite{yu2024visrag} to retrieve relevant image crops, guided by query-image semantic similarity and MLLM evaluation, while preserving the relative spatial layout of retrieved regions. This retrieval-based approach significantly outperforms zooming-based methods, especially in complex multi-object scenes.
% Inspired by the success of retrieval-augmented generation in Long-context LLMs ~\cite{jin2024long}, a recent work Retrieval-Augmented Perception (RAP) ~\cite{wang2025retrieval} introduce it in HR image understanding. RAP design an adaptive tree-search strategy and employ a pre-trained vision RAG model ~\cite{yu2024visrag} to retrieve  relevant image crops based on semantic similarity with the user query as well as MLLM evalutation while keep their relative position. This retrieval-based method significantly outperforms zooming-based methods especially in multi-object scene.

Despite its promising results, retrieval-based approaches have several intrinsic limitations that severely undermine their accuracy and robustness in challenging real-world scenarios: \textbf{\textit{1) Object Fragmentation.}} Fixed grid partitioning often splits large objects across multiple patches, destroying the holistic structural integrity of objects and producing fragmented embeddings with unreliable, biased similarity scores. \textit{\textbf{2) Scale Sensitivity.}} Crop resolution is an intractable hyperparameter for practical deployment. Overly large patches introduce excessive background noise that dilutes fine-grained target semantic signals, while excessively small crops further exacerbate the object fragmentation problem.
\textbf{\textit{3) Background Interference.}}  In HR images with textured, cluttered or crowded backgrounds, background regions frequently exhibit spurious high feature similarity to the query (e.g., from repetitive textures or shared visual statistics), generating false positives that mislead downstream MLLM inference. These failure modes collectively undermine the reliability of retrieval-augmented perception framework in complex HR image understanding tasks.

To address these challenges, we propose the \textbf{\textit{Multi-resolution Retrieval-Detection (MRD)}} framework, a multi-scale complementary paradigm for HR image understanding, as illustrated in Fig.~\ref{fig:mrd_short}. MRD systematically boosts retrieval accuracy and robustness via joint modeling of local semantic integrity and global spatial grounding within a unified architecture.
At the local scale, we introduce the \textit{Multi-resolution Semantic Fusion} module to adapt to target objects of diverse scales. Unlike conventional fixed patch-scale strategies, we compute semantic similarity across multiple proportional resolutions and enforce cross-scale consistency via fusion. This design effectively calibrates semantic bias, alleviates object fragmentation, and preserves local semantic integrity for precise target identification. At the global scale, we develop an \textit{Open-vocabulary Detector Enhancement} module for explicit global spatial grounding. As semantic similarity matching alone lacks sufficient spatial awareness for accurate target localization, we integrate the state-of-the-art open-vocabulary detector LLMDet~\cite{fu2025llmdet} as a robust global prior. Leveraging the in-context learning capability of LLMs, we extract target concepts from user queries to guide the generation of a target-aware confidence map, effectively suppressing false-positive background interference.
Finally, we augment multi-resolution semantic similarity with object detection confidence. This synergistic fusion amplifies responses of genuine target regions, enabling faster and more accurate relevant region localization during retrieval, and in turn guides MLLMs toward more reliable inference. 
Overall, MRD offers novel insights into joint modeling of local semantic integrity and global spatial grounding, advancing the frontier of HR image understanding. Extensive experiments demonstrate our superior performance and strong generalization across multiple mainstream benchmarks and diverse MLLM backbones. Our key contributions are summarized as follows:

\begin{itemize}\item We propose \textbf{\textit{MRD}}, a training-free unified multi-scale framework for high-resolution image understanding, which improves retrieval accuracy via jointly modeling local semantic integrity and global spatial grounding.
\item We design two core innovations for MRD: multi-resolution semantic fusion to address single-scale retrieval limitations, and open-vocabulary detector enhancement to complement semantic similarity for precise spatial localization and background suppression.
\item Comprehensive experiments across diverse benchmarks and MLLMs validate the superior performance and robustness of our method, providing new insights for advancing high-resolution image understanding.\end{itemize}
\section{Related Works}
\subsection{Multimodal Large Language Models}
\label{MLLMs}
Multimodal large language models have emerged as powerful foundation models, driving rapid progress in cross-modal understanding and generation across diverse vision-language tasks~\cite{zhang2024mm, yin2024survey}. A canonical MLLM architecture adopts a pre-trained vision encoder to map input images into visual tokens~\cite{radford2021learning, zhai2023sigmoid}, which are then projected into the embedding space of the backbone Large Language Model (LLM) via a multimodal connector, and jointly processed with textual inputs~\cite{dai2023instructblip, liu2024improved, liu2023visual}. Equipped with robust cross-modal alignment and emergent reasoning capabilities, MLLMs have become the core building block for a wide spectrum of multimodal applications, spanning medical image analysis~\cite{li2025towards, zambrano2025clinically}, video understanding~\cite{wang2024videoagent, wang2025videotree}, autonomous driving~\cite{hwangemma, xing2025openemma}, personalized assistants~\cite{nguyen2024yo, an2025unictokens}, and robotic manipulation~\cite{liu2024robomamba, li2024manipllm}. Despite these impressive advances, enabling MLLMs to achieve efficient, fine-grained understanding of high-resolution images remains a critical open challenge.
% MLLMs have emerged as powerful foundation models, driving rapid progress in cross-modal understanding and generation across diverse vision-language tasks~\cite{zhang2024mm, yin2024survey}. A canonical MLLM architecture adopts a pre-trained vision encoder to map input images into visual tokens~\cite{radford2021learning, zhai2023sigmoid}, which are then projected into the embedding space of the backbone LLM via a multimodal connector, and jointly processed with textual inputs~\cite{dai2023instructblip, liu2024improved, liu2023visual}. Equipped with robust cross-modal alignment and emergent reasoning capabilities, MLLMs have become the core building block for a wide spectrum of multimodal applications, spanning medical image analysis~\cite{li2025towards, zambrano2025clinically}, video understanding~\cite{wang2024videoagent, wang2025videotree}, autonomous driving~\cite{hwangemma, xing2025openemma}, personalized assistants~\cite{nguyen2024yo, an2025unictokens}, and robotic manipulation~\cite{liu2024robomamba, li2024manipllm}. Despite these impressive advances, enabling MLLMs to achieve efficient, fine-grained understanding of high-resolution images remains a critical open challenge.

\subsection{High-Resolution Image Understanding}
Extensive research efforts have been dedicated to advancing the HR image understanding capabilities of MLLMs, with mainstream works falling into several directions. The first line of work focuses on architectural improvement  for MLLMs. These methods either design HR-specialized visual encoders~\cite{lu2024deepseek, li2025mini,luo2024feast, ge2024convllava}, or adopt dynamic resolution strategies for patch-based encoding~\cite{zhang2024beyond,li2024llava, dong2024internlm, zhu2025internvl3,chen2024internvl}, to preserve fine-grained visual details in HR inputs. 
% However, they inherently suffer from severe visual redundancy and fragmentation, especially for ultra-high-resolution images. 
Another dominant paradigm is the localize-and-zoom-in framework. Methods under this paradigm leverage SFT~\cite{shao2024visual,wu2024v} or RL~\cite{zhang2025thyme, zheng2025deepeyes, li2025look, wang2025pixel} to train MLLMs to actively locate relevant regions and filter redundant background information. Despite their intuitive design, these methods suffer from prohibitive training costs, long convergence cycles, and poor cross-architecture transferability, which severely limit their scalability and real-world deployment~\cite{chu2025sft, yue2025does}.

Alternatively, a large body of training-free methods also follow the localize-and-zoom-in paradigm to enhance the HR understanding capability of off-the-shelf MLLMs. Works like TextCoT~\cite{luan2024textcot} and Zoom-Refine~\cite{yu2025zoom} directly leverage MLLMs’ inherent reasoning to locate target regions, while ZoomEye~\cite{shen2025zoomeye} and Dyfo~\cite{li2025dyfo} adopt tree-search to progressively narrow down relevant regions. Though effective on single-object tasks, they frequently miss targets in complex multi-object scenarios. Another branch of training-free works~\cite{zhang2025mllms, zhong2025focus, liu2025hide} uses MLLMs’ internal attention maps to guide region localization, but their performance heavily relies on the capacity of the pre-trained MLLM backbone. Recently, Retrieval-Augmented Perception~\cite{wang2025retrieval} integrates visual retrieval-augmented generation with tree-search to retrieve semantically relevant image patches, achieving state-of-the-art performance on both single and multi-object HR understanding tasks. Nevertheless, it still suffers from inherent limitations including object fragmentation and background interference.

\section{Preliminary}
\label{sec:Preliminary}
In this section, we first conduct an in-depth analysis of the failure cases of retrieval-based method to elucidate the inherent limitations of these approach. We then present an investigation into the correlation between the resolution of image crops and the performance of MLLMs. Building on the key insights derived from these empirical analyses, we introduce our \textbf{\textit{MRD}} framework in Sec.~\ref{sec:Method}.
% We first conduct an analysis of the relationship between the resolution of image crops and the performance of MLLMs in \autoref{subsec:Impact of the Resolution}. The experimental results indicate that using different resolution has a significant impact on MLLMs to analyze HR images. Objects of different sizes are suitable for different resolutions. Inspired by this we propose the MRD framework.
\subsection{Semantic Similarity}
\label{subsec:semantic similarity}
This section presents the pipeline for integrating retrieval-augmented generation into MLLMs to calculate the semantic similarity between the query embedding and image crops from HR images. Given an HR image, we first partition it collection of image crops, denoted as 
$P = \{ p_1, p_2, \ldots, p_n \}$, where $n$ is the total number of image crops. Following VisRAG~\cite{yu2024visrag}, the query and each image crop are encoded respectively using the text and image encoders of a Vision-Language Model (VLM), producing a sequence of hidden representations. The similarity score $s(q, p_i)$ between the query embedding and the $i$-th image crop embeddings can be calculated by the cosine similarity:
\begin{equation}
  s(q, p_i) = \frac{1}{2} \cdot (1 + \frac{q\cdot p_i}{\Vert q\Vert \cdot \Vert p_i\Vert })
  \label{eq:cos sim}
\end{equation}

Based on these scores, the $K$ most relevant image crops are selected and provided to the MLLM to support detailed understanding of the high-resolution input.

\begin{figure*}
  \centering
  \includegraphics[width=0.93\linewidth]{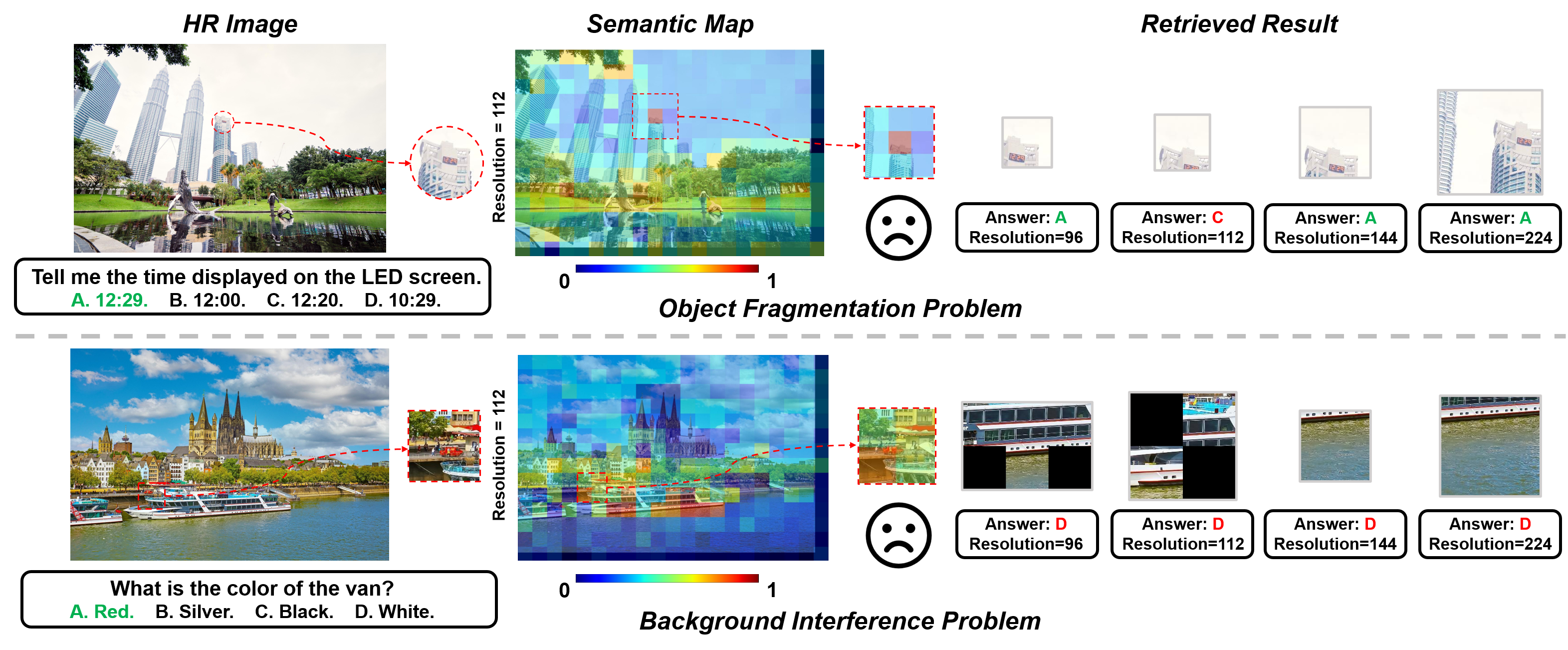}
    \caption{Object fragmentation and background interference problems in retrieval-based methods}
  \label{fig:limitations}
\end{figure*}

\subsection{Limitations of Retrieval-Based Methods}
\label{subsec:Limitations}
To systematically characterize the inherent limitations of retrieval-based HR image understanding methods, we conduct a fine-grained case-by-case analysis of failure cases produced by the RAP~\cite{wang2025retrieval} method on the \textit{V* Bench}~\cite{wu2024v} with LLaVA-ov~\cite{li2024llava}. We manually categorize the root causes of each failure based on the MLLM’s reasoning outputs, retrieved results, and semantic similarity scores, and define three core failure modes: \textbf{\textit{1) Object Fragmentation (FRAG)}}: Semantic bias arises from target object fragmentation, where the retrieval result only retains partial crops of the target object (Fig.~\ref{fig:limitations}, top). \textbf{\textit{2) Background Interference (BG)}}: Retrieved results are either dominated by or exclusively consist of high-similarity false-positive crops from irrelevant background regions (Fig.~\ref{fig:limitations}, bottom); (3) Other causes (e.g., inherent limitations of the MLLM itself). Notably, a single failure case can be attributed to multiple failure modes simultaneously. We quantify the impact of each failure mode by calculating the proportion of affected samples across the full dataset, with results summarized in the first row of Tab.~\ref{tab:Ablation}.  For the RAP framework , \textbf{\textit{FRAG}} and \textbf{\textit{BG}} errors affect 10.7\% and 8.9\% of the full dataset, accounting for 65.2\% and 54.3\% of all failure cases, respectively. Most notably, 89.6\% of all failure cases are associated with either \textbf{\textit{FRAG}} or \textbf{\textit{BG}} errors, confirming that these two failure modes are the dominant bottlenecks restricting the performance of retrieval-based HR image understanding methods.

\subsection{Impact of Image Crop Resolution}
\label{subsec:Impact of the Resolution}
\begin{figure}[t]
  \centering
   \includegraphics[width=1\linewidth]{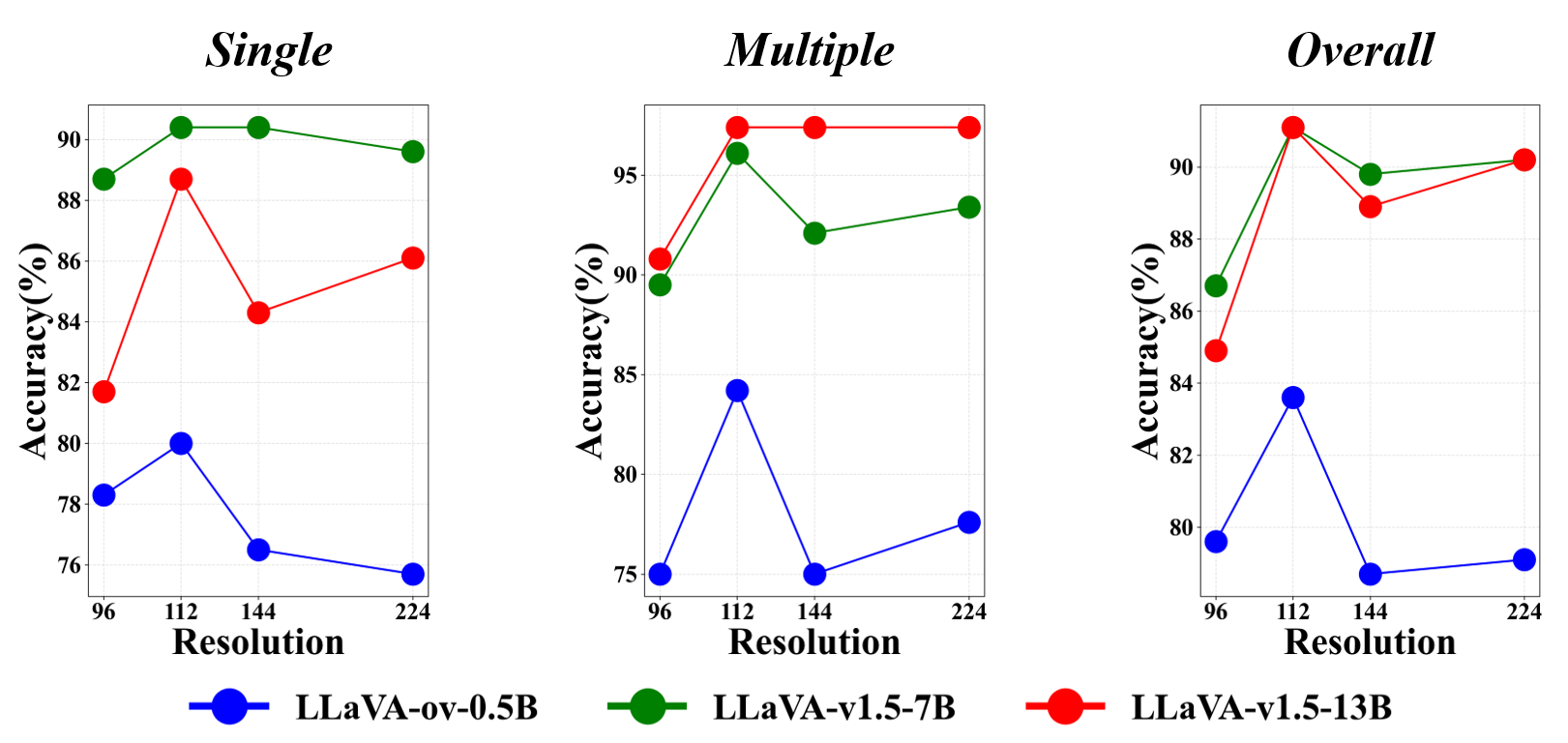}

   \caption{ The impact of the resolution of retrieved image crops on MLLMs performance.}
   \label{fig:resolution impact}
\end{figure}

To validate the influence of crop resolution, we conduct comprehensive experiments on the \textit{V* Bench} with the RAP method, adopting image crops of varying resolutions across a series of mainstream MLLMs; the results are summarized in Fig.~\ref{fig:resolution impact}. We observe that the selection of crop resolution imposes a pronounced impact on MLLMs’ performance on HR image understanding. Specifically, while a crop resolution of 112 yields the best performance on both single and multi-object tasks, it suffers from degraded recognition accuracy compared to alternative resolutions in several cases, which is caused by the object fragmentation problem (Fig.~\ref{fig:limitations} top). This key observation motivates us that partitioning HR images with adaptive crop resolutions, customized for target objects with diverse sizes and spatial locations, can be a more effective paradigm for HR visual understanding.

\section{Method}
\label{sec:Method}
Building on aforementioned the analysis and findings in Sec.~\ref{sec:Preliminary}, we propose a novel framework,  \textbf{\textit{Multi-resolution Retrieval Detection (MRD)}}. The core  of \textbf{\textit{MRD}} lies in its multi-scale multi-resolution design,  which achieves more precise target localization, suppresses background interference while preserving target semantic integrity to improve MLLMs perceptual performance for robust HR image understanding. Driven by insights in  Sec.~\ref{subsec:Impact of the Resolution}, we first design a simple yet effective \textit{Multi-resolution Semantic Fusion} module, which adapts to target objects of varying sizes and maintains the semantic integrity of objects at the local scale. On this basis, we propose an \textit{Open-vocabulary Detector Enhancement} module, which incorporates an open-vocabulary detection model to traverse the entire HR image via a sliding window approach at the global scale, enabling more accurate and direct localization of target regions while eliminating irrelevant background distractions. The synergistic integration of these two modules enables more accurate and efficient localization of target regions during the retrieval process, while simultaneously eliminating interference from irrelevant background. We elaborate on the complete algorithmic details and technical implementation of each component in the subsequent sections.
% multi-resolution approach at different scales to better localize regions containing target objects. This enables subsequent search processes to more easily identify image crops corresponding to the target objects, eliminating irrelevant distractions and enhancing the perceptual understanding of HR images by MLLMs. Based on the findings in Sec. \ref{subsec:Impact of the Resolution}, we argue that using different resolutions for semantic similarity computation is more suitable for objects of varying sizes and locations. Inspired by this idea, we first introduce a simple yet effective Multi-resolution Semantic Fusion method, which computes semantic similarity maps at different resolutions on a local scale and performs consistency-based fusion to refine the semantic similarity and improve its accuracy. To more directly localize target objects, we incorporate an Open-vocabulary object detection model that traverses the entire HR image globally using a sliding window approach, generating confidence scores for regions containing target objects. 
% Finally, by integrating the detection confidence scores with the multi-resolution semantic similarity maps, our method not only improves localization of target regions but also distinguishes fine-grained differences among crops in these regions, thereby assisting subsequent search processes in more accurately identifying key areas. The following sections will provide detailed explanations of each component.

\begin{figure*}
  \centering
  \includegraphics[width=0.92\linewidth]{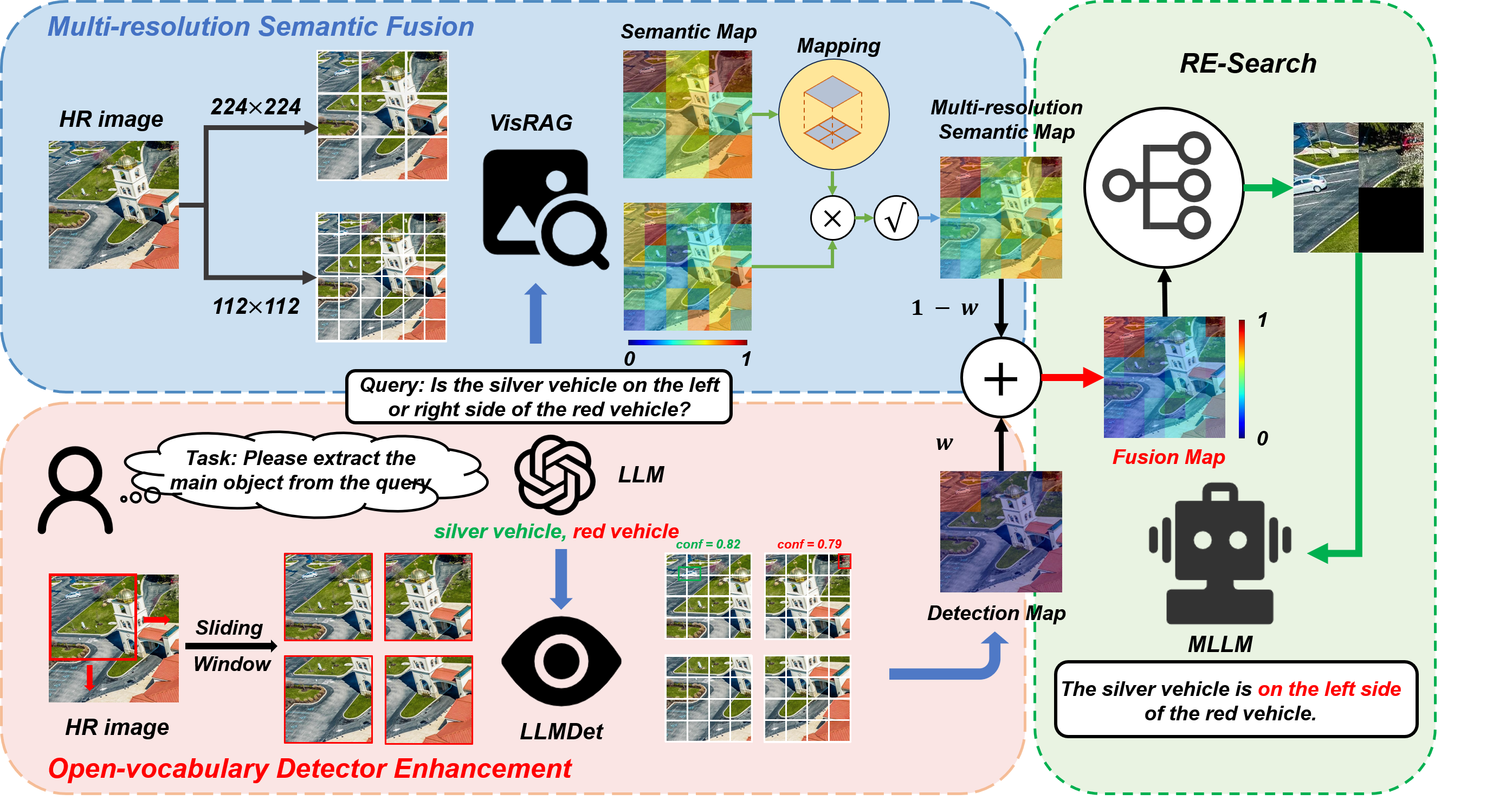}
    \caption{Overview of the proposed $\textit{\textbf{MRD}}$ framework. The framework generates a multi-resolution semantic similarity map via VisRAG from multi-scale crops, and a global detection confidence map via the open-vocabulary detector LLMDet with sliding-window target localization. The two maps are linearly fused, with the fused map guiding target-aware crop selection for subsequent search.}
  \label{fig:overview}
\end{figure*}

\subsection{Multi-resolution Semantic Fusion}
\label{subsec:Multi-resolution Semantic Fusion}
% In Sec. \ref{subsec:Impact of the Resolution}, we observe that image crops of different resolutions are suitable for objects of varying sizes and locations in different cases. Compared to the semantic similarity map obtained using a single resolution, those derived from multiple resolutions exhibit respective advantages. Therefore, we first propose a Multi-Resolution Semantic Fusion method. As shown in the top part of Fig. \ref{fig:overview}, we partition the HR image using proportional resolutions, with the low resolution set to $l$ and the high resolution set to $\hat{l}$, where $\hat{l} = k \cdot l$. The set of image crops at high resolution is denoted as $\hat{P} = \{ \hat{p}_1, \hat{p}_2, \ldots, \hat{p}_m \}$, and at low resolution as $P = \{p_1, p_2, \ldots, p_n\}$. Due to the proportional relationship between high and low resolutions, we have $n=k^2 \cdot m$, and each high-resolution crop $\hat{p_i}$ corresponds to $k^2$ low-resolution crops:
In Sec.~\ref{subsec:Impact of the Resolution}, we empirically observe that image crops at different resolutions are suited for capturing objects of varying scales across scenarios. Compared to semantic similarity maps generated from a single resolution, maps derived from complementary resolutions exhibit distinct and mutually beneficial strengths. To this end, we introduce a \textit{Multi-resolution Semantic Fusion} module to aggregate cross-resolution semantic information, as illustrated in the upper branch of Fig.~\ref{fig:overview}.
Specifically, we split the input high-resolution image into two sets of image crops with proportional resolutions. We denote the base resolution of the low-resolution (LR) crop set as $l$, and the resolution of the HR crop set as $\hat{l} = k \cdot l$, where $k$ is a positive integer scaling factor. Let $\hat{P} = \{ \hat{p}_1, \hat{p}_2, \ldots, \hat{p}_m \}$ denote the set of HR crops and  $P = \{p_1, p_2, \ldots, p_n\}$ denote the set of LR crops. Owing to the proportional scaling between the two resolutions, we have 
$n=k^2 \cdot m$ meaning each HR crop $\hat{p_i}$ spatially corresponds to exactly $k^2$ LR crops arranged in a 
$ k\times k$ grid:

% \begin{equation}
% \hat{p_i} =
% \begin{bmatrix}
% \tilde{p}_{i,1} & \tilde{p}_{i,2} & \cdots & \tilde{p}_{i,k} \\
% \tilde{p}_{i,(k+1)} & \tilde{p}_{i,(k+2)} & \cdots & \tilde{p}_{i,(2k)} \\
% \vdots & \vdots & \ddots & \vdots \\
% \tilde{p}_{i,(k(k-1)+1)} & \tilde{p}_{i,(k(k-1)+2)} & \cdots & \tilde{p}_{i,k^2}
% \end{bmatrix}
% \label{eq:Pi_matrix}
% \end{equation}

\begin{equation}
\hat{p_i} =
\begin{bmatrix}
\tilde{p}_{i,1} & \cdots & \tilde{p}_{i,k} \\
\vdots  & \ddots & \vdots \\
\tilde{p}_{i,(k(k-1)+1)} & \cdots & \tilde{p}_{i,k^2}
\end{bmatrix}
\label{eq:Pi_matrix}
\end{equation}

where each element $\tilde{p}_{i,t}$ corresponding to a unique LR crop $p_{j} \in P$, $i \in \{1, 2, \dots, m\}, \quad t \in \{1, 2, \dots, k^2\}$. 

Using the cosine similarity function defined in Eq.~\eqref{eq:cos sim}, we compute the semantic similarity between the query embedding and crops embedding, yielding two sets of similarity scores: $\hat{S} = \{ \hat{s}_1, \hat{s}_2, \ldots, \hat{s}_m \}$ for the HR crop set, and $S = \{s_1, s_2, \ldots, s_n\}$ for the LR crop set, formalized as:

\begin{equation}
\hat{s}_i = s(f(q), g(\hat{p}_i)), \quad s_j = s(f(q), g(p_j))
\label{eq:semantic map}
\end{equation}

where $f(\cdot)$ and $g(\cdot)$ denote the text encoder for the input query and the image encoder for image crops, respectively, with $i \in \{1, 2, \dots, m\}$ and $j \in \{1, 2, \dots, n\}$. 

Leveraging the spatial correspondence between HR and LR crops defined in Eq.~\eqref{eq:Pi_matrix}, we project each HR similarity score $\hat{s}_i$ to its corresponding $k^2$ LR crop positions, replicating the score across the associated $k\times k$ LR grid. This projection operation is written as:

\begin{equation}\tilde{S} = H(\hat{S})\label{eq:high to low}\end{equation}

where $\tilde{S}=\{\tilde{s}_1, \tilde{s}_2, \ldots, \tilde{s}_n\}$  is the projected HR similarity set, which shares the exact spatial dimensions and element count as the LR similarity set $S$, ensuring perfect crop-wise alignment between the two. With the spatially aligned similarity sets $\tilde{S}$ and $S$, 
we perform element-wise consistency fusion to generate the final multi-resolution semantic similarity scores 
$S^{m} = \{s^{m}_1, s^{m}_2, \ldots, s^{m}_n\}$, computed as:
\begin{equation}
s^{\text{m}}_t = \sqrt{\tilde{s}_t \cdot s_t},\ \ \ t \in {1, 2, \dots, n}
\end{equation}

Finally, we reshape the 1D fused similarity set $S^{m}$ into a 2D semantic similarity map $s^{\text{m}}(i,j)$ of size $H \times W$, where $H$ and $W$ denote the height and width of the LR crop grid respectively. 
Fusing cross-resolution semantic similarities enables correction of LR similarity degradation caused by complete objects being split across multiple LR crops. Our method reinforces consistent similarity responses across all parts of the target object, preserving its semantic integrity in subsequent retrieval steps.

\subsection{Open-vocabulary Detector Enhancement}
% To mitigate background interference in the multi-resolution semantic similarity map, we propose an \textit{Open-vocabulary Detector Enhancement} module, which introduces explicit object-level localization constraints via the open-vocabulary detector LLMDet with a crop-aligned sliding-window strategy. First, we leverage in-context learning with a LLM to extract core target entities from the input query, which serve as open-vocabulary detection categories for LLMDet. To align perfectly with the semantic similarity map generated by our multi-resolution fusion module, we assign detection confidence scores of target bounding boxes to their corresponding image crops, producing a detection confidence map that indicates the likelihood of target presence in each crop. This map provides explicit, spatially precise localization guidance complementary to the fine-grained semantic similarity map. We detail the pipeline below.
To alleviate background interference in the multi-resolution semantic similarity map, we propose an \textit{Open-vocabulary Detector Enhancement} module. This module introduces explicit object-level localization constraints using the open-vocabulary detector LLMDet, paired with a crop-aligned sliding-window strategy. First, we harness the in-context learning capability of LLMs to extract core target entities from the input query, which serve as the target categories for the open-vocabulary detector. To achieve precise alignment with the semantic similarity map generated by our multi-resolution fusion module, we map the detection confidence scores of target bounding boxes to their corresponding image crops, yielding a detection confidence map that quantifies the target presence probability for each crop. This map delivers explicit, spatially accurate localization guidance that is inherently complementary to our fine-grained semantic similarity map. The details are discussed below.
% To eliminate the interference of irrelevant background and achieve more direct and accurate global target localization, we propose \textit{Open-vocabulary Detector Enhancement} module which employ an advanced open-vocabulary object detection model-LLMDet directly localizing regions containing target objects of interest through sliding- window strategy. First, we employ in-context learning with a Large Language Model (LLM) to extract the primary target objects from the query, which serve as the target categories for LLMDet. Due to the extremely high resolution of HR images in datasets such as HR-Bench, we adopt a sliding window strategy for object localization. To align with the semantic similarity map derived from image crops, we assign the detection confidence scores of the target bounding boxes to their corresponding image crops, thereby generating a detection map that reflects the confidence of target object presence in each crop. This detection map offers a more intuitive localization representation compared to the semantic similarity map. In the following, we provide a detailed introduction to the proposed method.
% \vspace{0.5em}

\vspace{0.3em}
\noindent
\textbf{Object Extraction.}
To support open-vocabulary detection without pre-defined category limits, we first use an LLM to dynamically extract target object entities from free-form textual queries. Given an input query $Q$, we adopt in-context learning to extract primary target entities, which form the detection category set for our framework. Formally, the extraction process is defined as:

\begin{equation}
O = {\text{LLM}}(\mathcal{P}{\text{system}}, \mathcal{E}{\text{examples}}, Q)
\end{equation}
where $O$ denotes the set of extracted target objects, 
 $\mathcal{P}{\text{system}}$ is the system prompt specifying extraction rules, and $\mathcal{E}{\text{examples}}$ contains few-shot demonstration examples for in-context learning.

\vspace{0.3em}
\noindent
\textbf{Sliding-window Object Detection.}
To handle ultra-high-resolution images that exceed the detector’s optimal input size, we adopt a sliding-window detection strategy  which is strictly spatially aligned with the $ H \times W $ non-overlapping crop grid used in our multi-resolution semantic fusion module  (with total crop count $  n = H \times W $).

Specifically, a sliding window of size $ h \times w $ crops (where $ h < H $ and $ w < W $)  traverses the entire HR image with a pre-defined stride, yielding $T$ sliding windows denoted as $\Omega
 = \{w_1, w_2, ..., w_T\}$. For each sliding window $w_t$ we use LLMDet to perform open-vocabulary detection with the extracted target objects  $O$. The detector outputs a set of bounding boxes $ \mathcal{B}_{t} = \{b_1, b_2, \dots, b_{K_t}\} $ and corresponding confidence scores $c_k$, where $c_k$ indicates the confidence that bounding box $b_k$ contains the target object.

We first apply a confidence threshold $ \tau $ to filter out low-quality detections:
\begin{equation}
\mathcal{B}_{t}^{\text{filter}} = \{ b_k \in \mathcal{B}_{t} \mid c_k > \tau \}
\end{equation}
Subsequently, we generate a window detection confidence map $ \mathbf{c}^w_{t} \in \mathbb{R}^{h \times w} $ for each sliding window. For a crop at local coordinate $ (p, q) $ within the window, we assign the maximum confidence score of all filtered bounding boxes that cover this crop; the score is set to 0 if no valid box covers the crop:

\begin{equation}
c^{\text{w}}_{t}(p, q) = \max_{b_k \in \mathcal{B}_{t}^{\text{filter}}} \{ c_k \cdot \mathbb{I}[(p, q) \in b_k] \}
\end{equation}
where $ \mathbb{I}[\cdot] $ is the indicator function, which equals 1 if the crop at $ (p, q) $ lies within bounding box $ b_k $, and 0 otherwise.

To aggregate local window into a global unified detection confidence map $ \mathbf{c}^{\text{g}} \in \mathbb{R}^{H \times W} $ (aligned with the semantic similarity map), we adopt an average fusion strategy to mitigate boundary artifacts from overlapping windows and smooth single-window detection noise. For a crop at global coordinate $(i,j)$ in the HR image grid, we denote $\mathcal{T}_{i,j}$ as the set of sliding windows that contain this crop, and $(t_i, t_j)$ as its corresponding local coordinate in the 
$t$-th sliding window $w_t$. The final global confidence score is computed as the average of all valid local scores assigned to this crop:
\begin{equation}
c^{\text{g}}(i, j) = \frac{1}{|\mathcal{T}_{i,j}|} \sum_{t \in \mathcal{T}_{i,j}} c^{\text{w}}_{t}(t_i, t_j)
\end{equation}
where $i \in \{1, 2, \dots, H\}$ and $j \in \{1, 2, \dots, W\}$. 

\vspace{0.3em}
\noindent
\textbf{Semantic-Detection Fusion.}
The global detection confidence map provides explicit target localization and strong background suppression, but lacks the ability to capture fine-grained semantic differences within the target object. In contrast, our multi-resolution semantic similarity map encodes fine-grained query-image matching, but is vulnerable to background clutter and boundary ambiguity. To leverage the complementary strengths of the two branches, we fuse them via a linear combination to generate the final fused similarity map:

\begin{equation}
s^{\text{f}}(i,j)= (1-w)\cdot s^\text{m}(i,j) + w \cdot c^\text{g}(i,j)
\end{equation}
where $w$ is a balancing weight. This synergistic fusion strategy simultaneously achieves precise target localization and full semantic integrity preservation, which in turn enables more efficient and accurate extraction and retrieval of key regions during the subsequent search process. The detailed description of the follow-up \textit{Retrieved-Exploration Search} pipeline is provided in~\cite{wang2025retrieval}.

\section{Experiments}
\subsection{Experimental Settings}
% \vspace{0.3em}
\noindent
\textbf{Benchmarks.} We evaluate our \textit{MRD} on two HR image benchmarks: $V^*$ \textit{Bench}~\cite{wu2024v} and \textit{HR-Bench}~\cite{wang2025divide} with \textit{4K} or \textit{8K} resolution. Both cover single-object and multi-object understanding tasks for HR images.

\vspace{0.3em}
\noindent
\textbf{Baseline Methods.} We evaluate our  \textit{MRD} against three groups of baselines on HR benchmarks: 1) mainstream open-source MLLMs, including LLaVA-v1.6 series~\cite{liu2024llavanext}, LLaVA-HR-X~\cite{DBLP:conf/iclr/LuoZ0ZSJ25}, InternVL-1.5~\cite{DBLP:journals/chinaf/ChenWTYGCTHLMMWDYGHSJXW24} and Yi-VL ~\cite{DBLP:journals/corr/abs-2403-04652}; 2) closed-source MLLMs GPT-4o~\cite{hurst2024gpt} and Qwen-VL-max~\cite{bai2023qwenvlversatilevisionlanguagemodel} as upper-bound references; 3) SOTA high-resolution specialized methods Zoom Eye~\cite{shen2025zoomeye} and RAP~\cite{wang2025retrieval}, which share the same LLaVA-v1.5-7B~\cite{liu2024improved} and LLaVA-ov-0.5B~\cite{li2024llava} backbones with our approach for fair comparison.
\begin{table*}[!t]
    \caption{Comparison of our \textbf{\textit{MRD}} with existing works on high-resolution benchmarks}
    \centering
    \small % 替换原\small，缩小字号（需更小可换\scriptsize）
    \renewcommand{\arraystretch}{1.0} % 可选：压缩行高，默认值为1
    \begin{tabular}{lccccccccc}
    \toprule
        \multirow{2}{*}{\textbf{Method}} & \multicolumn{3}{c}{\textbf{\textit{V* Bench}}} & \multicolumn{3}{c}{\textbf{\textit{HR-Bench 4K}}} & \multicolumn{3}{c}{\textbf{\textit{HR-Bench 8K}}} \\
    \cmidrule(r){2-4}\cmidrule(r){5-7}\cmidrule(r){8-10}
        ~ & \textbf{\textit{Attribute}} & \textbf{\textit{Spatial}} & \textbf{\textit{Overall}} & \textbf{\textit{FSP}} & \textbf{\textit{FCP}} & \textbf{\textit{Overall}} & \textbf{\textit{FSP}} & \textbf{\textit{FCP}} & \textbf{\textit{Overall}} \\
    \midrule
        \multicolumn{10}{c}{\textit{Open-source MLLMs}} \\
    \midrule
        LLaVA-v1.6-7B~\cite{liu2024llavanext} & 60.9 & 63.2 & 61.8 & 49.0 & 46.8 & 47.9 & 37.3 & 44.3 & 40.8 \\
        LLaVA-v1.6-13B~\cite{liu2024llavanext} & 60.0 & 64.5 & 61.8 & 49.8 & 41.3 & 45.5 & 38.0 & 38.3 & 38.1 \\
        LLaVA-v1.6-34B~\cite{liu2024llavanext} & - & - & - & 55.3 & 50.5 & 52.9 & 44.5 & 50.3 & 47.4 \\
        LLaVA-HR-X-13B~\cite{DBLP:conf/iclr/LuoZ0ZSJ25} & - & - & - & 61.3 & 46.0 & 53.6 & 49.5 & 44.3 & 46.9 \\
        LLaVA-HR-X-7B~\cite{DBLP:conf/iclr/LuoZ0ZSJ25} & 51.3 & 64.5 & 56.5 & 57.8 & 46.3 & 52.0 & 42.0 & 41.3 & 41.6 \\
        InternVl-1.5-26B~\cite{DBLP:journals/chinaf/ChenWTYGCTHLMMWDYGHSJXW24} & - & - & - & 69.5 & 51.8 & 60.6 & 69.3 & 48.5 & 57.9 \\
        Yi-VL-34B \cite{DBLP:journals/corr/abs-2403-04652} & - & - & - & 46.0 & 42.8 & 44.4 & 39.5 & 38.5 & 39.0 \\
    \midrule
        \multicolumn{10}{c}{\textit{Closed-source MLLMs}} \\
    \midrule
        GPT-4o~\cite{hurst2024gpt} & - & - & 66.0 & 70.0 & 48.0 & 59.0 & 62.0 & 49.0 & 55.5 \\
        Qwen-VL-max~\cite{bai2023qwenvlversatilevisionlanguagemodel} & - & - & - & 65.0 & \textbf{52.0} & 58.5 & 54.0 & \textbf{51.0} & 52.5 \\
    \midrule
        \multicolumn{10}{c}{\textit{Baselines and MRD}} \\
    \midrule
        LLaVA-v1.5-7B~\cite{liu2024improved} & 43.5 & 56.6 & 48.7 & 38.5 & 33.8 & 36.1 & 33.0 & 31.3 & 32.1 \\
        LLaVA-v1.5-7B-Zoom Eye~\cite{shen2025zoomeye} & 83.5 & 82.9 & 83.3 &  67.8 &  38.8 & 53.3 &  65.5 & 36.0 & 50.8 \\
        LLaVA-v1.5-7B-RAP~\cite{wang2025retrieval} & 90.4 & \textbf{96.1} & 91.1 & 73.8 & 40.5 & 57.1 & 72.3 & 35.3 & 53.8 \\
        \rowcolor{gray!15}
        \textbf{LLaVA-v1.5-7B-MRD} \textbf{\textit{(ours)}} & \textbf{97.4} & \textbf{96.1} & \textbf{95.6} & 76.8 & 42.7 & 59.9 & 73.0 & 36.5 & 54.8 \\
    \hline
    %         LLaVA-v1.5-13B \cite{liu2024llavanext} & 41.7 & 55.3 & 47.1 & 45.2 & 41.3 & 43.3 & 37.5 & 38.0 & 37.8 \\
    %     LLaVA-v1.5-7B-Zoom Eye \cite{shen2025zoomeye} & 87.8 & 81.6 & 85.3 &  73.0 &  43.3 & 58.1 & 67.3 & 45.5 & 56.4 \\
    %     LLaVA-v1.5-13B-RAP \cite{wang2025retrieval} & 89.6 & 90.8 & 89.8 & 74.3 & 46.0 & 60.1 & 69.8 & 38.2 & 54.0 \\
    %     \rowcolor{gray!15}
    %     \textbf{LLaVA-v1.5-13B-MRD} \textbf{\textit{(ours)}} & \textbf{97.4} & 94.8 & 95.1 & 76.0 & 46.0 & 61.0 & 75.3 & 39.7 & 57.5 \\
    % \hline
        LLaVA-ov-0.5B~\cite{li2024llava} & 63.5 & 64.5 & 63.9 & 63.5 & 39.5 & 51.5 & 47.3 & 38.3 & 42.8 \\
        LLaVA-ov-0.5B-Zoom Eye~\cite{shen2025zoomeye} &  85.2 &  73.7 &  80.6 & 75.5 & 39.8 & 57.6 &  68.5 & 38.3 & 53.4 \\
        LLaVA-ov-0.5B-RAP~\cite{wang2025retrieval} & 80.0 & 84.2 & 83.6 & 80.3 & 42.3 & 61.3 & \textbf{81.8} & 45.3 & 63.5 \\
        \rowcolor{gray!15}
        \textbf{LLaVA-ov-0.5B-MRD} \textbf{\textit{(ours)}} & 89.6 & 85.6 & 88.9 & \textbf{84.0} & 45.2 & \textbf{64.6} & \textbf{81.8} & 47.3 & \textbf{64.5} \\
    \bottomrule
    \end{tabular}
    \label{tab:overall_results}
\end{table*}

% \begin{figure*}[htbp]
%     \centering

%     % -------- 子图 1 --------
%     \begin{subfigure}[b]{0.75\linewidth}   % 控制宽度
        
%         \centering
%         \includegraphics[width=\linewidth]{fig/multi_res.png}
%         % \label{fig:Multi-resolution-Semantic-Fusion-Effect}
%     \end{subfigure}

%     % \vspace{-0.5cm}  % 控制上下子图间距

%     % -------- 子图 2 --------
%     \begin{subfigure}[b]{0.75\linewidth}
%         \centering
%         \includegraphics[width=\linewidth]{fig/OVD.png}
%         \label{fig:sub2}
%     \end{subfigure}
%     % \vspace{-0.9cm}  % 控制上下子图间距

%     \caption{Visualization of the Effects of Different Modules in MRD. Upper: Visualization of the Effects of the Multi-resolution Semantic Fusion Method. Lower: Visualization of the Effects of the Multi-resolution Semantic Fusion Method}
%     \label{fig:two_subfig_vertical}
% \end{figure*}

% \clearpage
% \vspace{4em}

\subsection{Main Results}
As shown in Tab.~\ref{tab:overall_results}, our proposed \textit{MRD} framework consistently achieves significant performance gains across all high-resolution benchmarks, sub-tasks, and model configurations, outperforming both vanilla MLLM baselines and state-of-the-art high-resolution perception methods.
Built on LLaVA-v1.5-7B, MRD delivers a 46.9\% absolute improvement in overall accuracy on \textit{V* Bench} (nearly doubling the baseline), along with up to 23.8\% and 22.7\% absolute overall gains on \textit{HR-Bench 4K} and \textit{HR-Bench 8K}, respectively. It consistently surpasses the SOTA RAP method across all datasets and model settings, with an average overall improvement of 2.8\%.
MRD brings stable performance boosts to both single and multi-object tasks, with particularly prominent gains on single-object tasks. For the single-object attribute recognition sub-task of \textit{V* Bench}, MRD achieves a 53.9\% absolute accuracy gain over the vanilla LLaVA-v1.5-7B baseline, and a 7.0\% gain over RAP. We attribute this advantage to the integrated detection module, which provides precise object localization for fine-grained visual understanding.
These results validate that MRD significantly enhances the high-resolution image perception and reasoning capabilities of MLLMs.

% As shown in Tab. \ref{tab:overall_results}, our proposed \textbf{\textit{MRD}} framework consistently delivers substantial performance gains across all high-resolution benchmarks, sub-tasks, and model configurations, outperforming both vanilla MLLM baselines and state-of-the-art high-resolution perception methods.
% Built on LLaVA-v1.5-7B, MRD achieves a remarkable 46.9\% absolute improvement in overall accuracy on V* Bench, nearly doubling the baseline performance. It also yields up to 23.8\% and 22.7\% absolute overall gains on \textit{HR-Bench 4K} and \textit{HR-Bench 8K}, respectively. Compared with the SOTA RAP method, MRD consistently surpasses it across all datasets and model settings, with an average overall improvement of 2.8\%.
% Notably, MRD brings consistent performance boosts to both single-object and multi-object tasks, with particularly prominent improvements on single-object tasks. For example, on the single-object attribute recognition sub-task of V* Bench, MRD achieves a 53.9\% absolute accuracy gain over the vanilla LLaVA-v1.5-7B baseline, and a 7.0\% gain over the SOTA RAP. We attribute this advantage to the integrated detection module, which provides precise object localization for fine-grained visual understanding.
% These results validate that MRD significantly enhances the high-resolution image perception and reasoning capabilities of MLLMs.

\begin{figure*}
  \centering
  \includegraphics[width=0.96\linewidth]{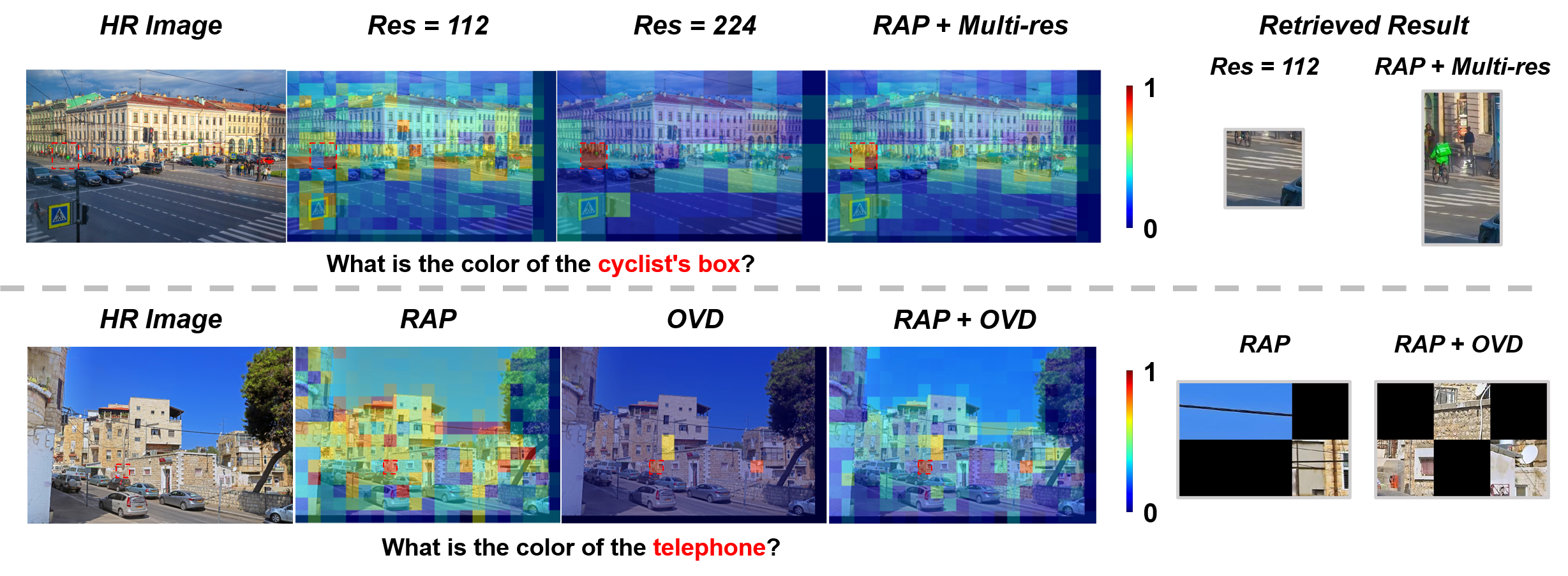}
    \caption{Visualization of the effects of proposed modules in \textbf{\textit{MRD}}. Upper: Visualization of the effects of the \textit{Multi-resolution Semantic Fusion} module. Lower: Visualization of the effects of the \textit{Open-vocabulary Detector Enhancement} module.}
  \label{fig:visual_modules}
\end{figure*}
% As shown in Tab. \ref{tab:overall_results},  compared with both the baseline MLLMs and previous baseline approaches, our proposed MRD framework consistently delivers substantial performance gains across all sub-tasks, datasets, and model configurations. The improvement is most pronounced on the $V^{*}$ \textit{Bench} using the LaVA-v1.5-7B model, where MRD achieves a remarkable 46.9\% absolute increase in accuracy—nearly doubling the original performance. Significant gains are also observed on HR-Bench 4K and HR-Bench 8K, with maximum improvements of 23.6\% and 22.8\%, respectively. 

% In comparison to the state-of-the-art baseline RAP, MRD achieves superior performance across all datasets and model settings, yielding an average improvement of 2.8\%.  When examining results across sub-task categories, MRD demonstrates particularly strong performance on single-object tasks. We attribute this advantage to the integration of a detection module, which provides more accurate localization for isolated objects.
% Overall, these results indicate that MRD markedly enhances the perception and understanding capabilities of MLLMs when operating on high-resolution images.

\subsection{Visualization of Module Effects}
To intuitively verify the effectiveness of our core modules, we visualize the confidence map and retrieved result using either \textit{Multi-resolution Semantic Fusion} module or the \textit{Open-vocabulary Detector Enhancement} module in Fig.~\ref{fig:visual_modules}.

The \textit{Multi-resolution Semantic Fusion} module obtains more accurate target information by integrating semantic similarity maps from different resolutions. As shown in the upper part of Fig.~\ref{fig:visual_modules}, this module leverages the high-resolution semantic similarity map to correct the semantic deviation caused by the splitting of the target object across multiple crops under single-resolution processing. This correction effectively alleviates the above semantic deviation and better preserves the integrity of the target object. 
% The visualization results demonstrate that the proposed module is effective for both single-object and multi-object tasks, and has better adaptability to objects of different sizes compared with the single-resolution setting.

For the \textit{Open-vocabulary Detector Enhancement} module, we introduce an open-vocabulary object detection model to achieve more accurate and direct localization of the target object at the global scale. As shown in the lower part of Fig.~\ref{fig:visual_modules}, the sliding-window detection results can effectively identify the location of the target object. By combining the detection results with the semantic similarity scores, our method amplifies the scores of the crops containing the target object, while suppressing the interference of false-positive crops with high semantic similarity and irrelevant background information. 

\subsection{Ablation Study} 
To better understand the contributions of different modules in our \textbf{\textit{MRD}} framework, we conduct comprehensive  quantitative ablation studies on the $V^*$ \textit{Bench} using the LLaVA-ov-0.5B model. As shown in Tab.~\ref{tab:Ablation}, the standalone OVD (first row) model achieves higher accuracy for single-object tasks via precise target localization and reduced background interference. However, its performance on multi-object tasks is inferior to RAP which may due to missing object in complex multiple object scene.  When equipping RAP with multi-resolution semantic fusion (second row), we observe consistent performance gains on both single and multi-object tasks, as it corrects semantic deviations, effectively mitigates object fragmentation, and preserves object completeness, making it capable of handling objects of varying sizes across diverse scenarios.

Fusing the semantic similarity map obtained from RAP with the detection confidence map from OVD (fourth row) further suppresses background interference and boosts single-object task performance. By further integrating multi-resolution semantic fusion, we achieve consistent and prominent improvements across both single- and multi-object tasks, verifying the effectiveness of our fusion strategy. The two core components are complementary and jointly necessary to address both background and fragmentation errors, ensuring consistent improvements across all settings. As a whole, our full MRD framework integrating all modules outperforms RAP by 5.3\% in overall accuracy, demonstrating the superiority of our MRD framework.
% ’s design for high-resolution image understanding in MLLMs.
% In summary, OVD improves single-object localization accuracy but risks object missing in multi-object scenarios, while multi-resolution semantic fusion calibrates semantic scores, maintains object integrity, and enhances MLLM performance on both tasks. 

\begin{table}
\caption{Ablation study of different module in \textbf{\textit{MRD}}.}
\centering
\huge
\renewcommand{\arraystretch}{1.2}

\resizebox{\columnwidth}{!}{
\begin{tabular}{lcccccc}
\toprule
\multirow{2}{*}[-0.3em]{\centering \textbf{Method}}
& \multicolumn{3}{c}{\textbf{\textit{Accuracy}}} 
& \multicolumn{3}{c}{\textbf{\textit{Error Rate}}} \\
\cmidrule(lr){2-4} \cmidrule(lr){5-7}
& \textbf{\textit{Attribute}} ($\uparrow$) 
& \textbf{\textit{Spatial}} ($\uparrow$) 
& \textbf{\textit{Overall}} ($\uparrow$) 
& \textbf{\textit{BG}} ($\downarrow$) 
& \textbf{\textit{FRAG}} ($\downarrow$) 
& \textbf{\textit{BG/FRAG}} ($\downarrow$) \\
\midrule
RAP 
& 80.0 
& 84.2 
& 83.6 
& 10.7 (64.2\%)
& 8.9 (54.3\%)
& 14.7 (89.6\%) \\
\cdashline{1-7}
OVD & 84.3 {\color{green!60!black}(+4.3)} 
& 82.9 {\color{red!70!black}(-1.3)} 
& 85.3 {\color{green!60!black}(+1.7)} 
& 5.7 {\color{green!60!black}(-46.7\%)} 
& 6.2 {\color{green!60!black}(-30.3\%)} 
& 9.3 {\color{green!60!black}(-36.7\%)} \\

RAP+Multi-Res 
& 83.5 {\color{green!60!black}(+3.5)} 
& \textbf{85.5}
{\color{green!60!black}(+1.3)} 
& 85.8 {\color{green!60!black}(+2.2)} 
& 6.7 {\color{green!60!black}(-37.4\%)} 
& 5.3 {\color{green!60!black}(-40.4\%)} 
& 9.7 {\color{green!60!black}(-34.0\%)} \\

RAP+OVD 
& 86.1 {\color{green!60!black}(+6.1)} 
& 84.2 (+0.0) 
& 86.7 {\color{green!60!black}(+3.1)} 
& 4.9 {\color{green!60!black}(-54.2\%)} 
& 5.8 {\color{green!60!black}(-34.8\%)} 
& 8.9 {\color{green!60!black}(-39.5\%)} \\
\rowcolor{gray!20}
All (MRD) 
& \textbf{89.6 {\color{green!60!black}(+9.6)}} 
& \textbf{85.5 {\color{green!60!black}(+1.3)}} 
& \textbf{88.9 {\color{green!60!black}(+5.3)}} 
& \textbf{4.0 {\color{green!60!black}(-62.6\%)}} 
& \textbf{4.4 {\color{green!60!black}(-50.6·\%)}} 
& \textbf{7.6 {\color{green!60!black}(-48.3\%)}} \\
\bottomrule
\end{tabular}
}
\label{tab:Ablation}
\end{table}

\subsection{Efficiency} 
We benchmark the average per-sample runtime and computational overhead of our \textit{\textbf{MRD}} against the RAP method on the \textit{HR-Bench 4K} dataset. Benefiting from our fused map that enables more accurate localization of target regions, we can drastically reduce the maximum number of search steps to 50 during the search phase while preserving full model performance. As shown in Tab.~\ref{tab:Efficiency}, although MRD introduces additional GPU memory consumption and computational cost due to multi-resolution fusion and sliding-window detection, it yields a remarkable acceleration of the search phase, which leading to a shorter overall runtime. 

% We compare the \textit{\textbf{per-sample average}} runtime and computational overhead of our \textit{\textbf{MRD}} against RAP on \textit{HR-Bench 4K}. Benefiting from our fused map that enables more accurate localization of target regions, we can drastically reduce the maximum number of search steps to 50 during the search phase while preserving full model performance. As shown in the Tab. \ref{tab:Efficiency}, although MRD introduces additional GPU memory and computation cost due to multi-resolution fusion and sliding-window detection, it significantly accelerates the search phase, leading to a shorter runtime. 

\section{Conclusion}
% In this work, we propose a novel training-free method, Multi-resolution Retrieval-Detection (MRD), to enhance the understanding of high-resolution images by MLLMs. MRD employs multi-resolution semantic similarity to correct single-resolution similarity maps, ensuring the integrity of target objects. Moreover, to localize target objects more accurately and directly, we introduce an OVD model that identifies object regions using a sliding-window approach. We demonstrate the effectiveness of MRD across multiple high-resolution benchmarks with different MLLMs, showing its superior performance in HR image understanding. We believe this work does indeed bring new insights into jointly modeling local semantic integrity and global spatial grounding in HR image understanding.
In this work, we present \textbf{\textit{Multi-resolution Retrieval-Detection (MRD)}}, a novel training-free method to boost MLLMs’ high-resolution  image understanding. MRD leverages multi-resolution semantic similarity to correct single-resolution similarity maps and preserve target object integrity, and introduces a sliding-window open-vocabulary detection (OVD) model for precise target localization. Extensive experiments on multiple HR benchmarks with various MLLMs validate MRD’s effectiveness and its superior HR image understanding performance. This work provides new insights into jointly modeling local semantic integrity and global spatial grounding for HR image understanding.
\begin{table}
\caption{Comparison of inference efficiency on \textit{HR-Bench4K}.}
\centering
\huge
\renewcommand{\arraystretch}{1.2}
\setlength{\tabcolsep}{4pt}

\resizebox{\columnwidth}{!}{
\begin{tabular}{lcccccc}
\toprule
\multirow{2}{*}{\textbf{Method}} & \multicolumn{5}{c}{\textbf{\textit{HR-Bench4K}}} \\
\cmidrule(lr){2-6}
 & \textbf{\textit{RAG (sec)}}  & \textbf{\textit{DET (sec)}} & \textbf{\textit{Search (sec)}} & \textbf{\textit{Total (sec)}} & \textbf{\textit{Max Mem (GB)}} \\
\midrule
RAP (ov-0.5B) & 9.6 & 0 & 55.6 & 67.0 & 18.9 \\
\rowcolor{gray!20}
MRD (ov-0.5B) & 14.7 {\color{red!70!black}(+53.1\%)} & 16.1 & 27.5 {\color{green!60!black}(-50.5\%)}  & 59.6 {\color{green!60!black}(-11.0\%)} & 21.2 {\color{red!70!black}(+12.2\%)} \\
\cdashline{1-6}
RAP (v1.5-7B) & 9.8 & 0 & 52.8 & 63.4 & 21.2 \\
\rowcolor{gray!20}
MRD (v1.5-7B) & 14.5 {\color{red!70!black}(+48.0\%)} & 15.8 & 15.2 {\color{green!60!black}(-71.2\%)} & 53.4 {\color{green!60!black}(-26.2\%)} & 23.4 {\color{red!70!black}(+10.4\%)}  \\
\bottomrule
\end{tabular}
}

\label{tab:Efficiency}
\end{table}

\section*{Acknowledgment}
This work is funded in part by the National Natural Science Foundation of China (Grant No. 62372480), GuangDong Basic and Applied Basic Research Foundation (2025A1515011361), Shenzhen Science and Technology Program (JCYJ20240813110459017).

\clearpage

{
    \small
    \bibliographystyle{ieeenat_fullname}
    \bibliography{main}

@String(ICLR = {Int. Conf. Learn. Represent.})

@String(AAAI = {AAAI})

@String(ICLR  = {ICLR})

@article{yin2024survey,
  title={A survey on multimodal large language models},
  author={Yin, Shukang and Fu, Chaoyou and Zhao, Sirui and Li, Ke and Sun, Xing and Xu, Tong and Chen, Enhong},
  journal={National Science Review},
  volume={11},
  number={12},
  pages={nwae403},
  year={2024},
  publisher={Oxford University Press}
}

@article{dai2023instructblip,
  title={Instructblip: Towards general-purpose vision-language models with instruction tuning},
  author={Dai, Wenliang and Li, Junnan and Li, Dongxu and Tiong, Anthony and Zhao, Junqi and Wang, Weisheng and Li, Boyang and Fung, Pascale N and Hoi, Steven},
  journal={Advances in neural information processing systems},
  volume={36},
  pages={49250--49267},
  year={2023}
}

@inproceedings{liu2024improved,
  title={Improved baselines with visual instruction tuning},
  author={Liu, Haotian and Li, Chunyuan and Li, Yuheng and Lee, Yong Jae},
  booktitle={Proceedings of the IEEE/CVF conference on computer vision and pattern recognition},
  pages={26296--26306},
  year={2024}
}

@inproceedings{zhai2023sigmoid,
  title={Sigmoid loss for language image pre-training},
  author={Zhai, Xiaohua and Mustafa, Basil and Kolesnikov, Alexander and Beyer, Lucas},
  booktitle={Proceedings of the IEEE/CVF international conference on computer vision},
  pages={11975--11986},
  year={2023}
}

@article{liu2023visual,
  title={Visual instruction tuning},
  author={Liu, Haotian and Li, Chunyuan and Wu, Qingyang and Lee, Yong Jae},
  journal={Advances in neural information processing systems},
  volume={36},
  pages={34892--34916},
  year={2023}
}

@inproceedings{radford2021learning,
  title={Learning transferable visual models from natural language supervision},
  author={Radford, Alec and Kim, Jong Wook and Hallacy, Chris and Ramesh, Aditya and Goh, Gabriel and Agarwal, Sandhini and Sastry, Girish and Askell, Amanda and Mishkin, Pamela and Clark, Jack and others},
  booktitle={International conference on machine learning},
  pages={8748--8763},
  year={2021},
  organization={PmLR}
}

@misc{liu2024llavanext,
  title={Llavanext: Improved reasoning, ocr, and world knowledge},
  author={Liu, Haotian and Li, Chunyuan and Li, Yuheng and Li, Bo and Zhang, Yuanhan and Shen, Sheng and Lee, Yong Jae},
  year={2024}
}

@article{bai2023qwen,
  title={Qwen technical report},
  author={Bai, Jinze and Bai, Shuai and Chu, Yunfei and Cui, Zeyu and Dang, Kai and Deng, Xiaodong and Fan, Yang and Ge, Wenbin and Han, Yu and Huang, Fei and others},
  journal={arXiv preprint arXiv:2309.16609},
  year={2023}
}

@article{zheng2025deepeyes,
  title={DeepEyes: Incentivizing" Thinking with Images" via Reinforcement Learning},
  author={Zheng, Ziwei and Yang, Michael and Hong, Jack and Zhao, Chenxiao and Xu, Guohai and Yang, Le and Shen, Chao and Yu, Xing},
  journal={arXiv preprint arXiv:2505.14362},
  year={2025}
}

@article{zhang2025mllms,
  title={Mllms know where to look: Training-free perception of small visual details with multimodal llms},
  author={Zhang, Jiarui and Khayatkhoei, Mahyar and Chhikara, Prateek and Ilievski, Filip},
  journal={arXiv preprint arXiv:2502.17422},
  year={2025}
}

@inproceedings{shen2025zoomeye,
  title={Zoomeye: Enhancing multimodal llms with human-like zooming capabilities through tree-based image exploration},
  author={Shen, Haozhan and Zhao, Kangjia and Zhao, Tiancheng and Xu, Ruochen and Zhang, Zilun and Zhu, Mingwei and Yin, Jianwei},
  booktitle={Proceedings of the 2025 Conference on Empirical Methods in Natural Language Processing},
  pages={6613--6629},
  year={2025}
}

@inproceedings{wu2024v,
  title={V?: Guided visual search as a core mechanism in multimodal llms},
  author={Wu, Penghao and Xie, Saining},
  booktitle={Proceedings of the IEEE/CVF Conference on Computer Vision and Pattern Recognition},
  pages={13084--13094},
  year={2024}
}

@inproceedings{li2025dyfo,
  title={Dyfo: A training-free dynamic focus visual search for enhancing lmms in fine-grained visual understanding},
  author={Li, Geng and Xu, Jinglin and Zhao, Yunzhen and Peng, Yuxin},
  booktitle={Proceedings of the Computer Vision and Pattern Recognition Conference},
  pages={9098--9108},
  year={2025}
}

@article{wang2025retrieval,
  title={Retrieval-augmented perception: High-resolution image perception meets visual rag},
  author={Wang, Wenbin and Jing, Yongcheng and Ding, Liang and Wang, Yingjie and Shen, Li and Luo, Yong and Du, Bo and Tao, Dacheng},
  journal={arXiv preprint arXiv:2503.01222},
  year={2025}
}

@inproceedings{wang2025divide,
  title={Divide, conquer and combine: A training-free framework for high-resolution image perception in multimodal large language models},
  author={Wang, Wenbin and Ding, Liang and Zeng, Minyan and Zhou, Xiabin and Shen, Li and Luo, Yong and Yu, Wei and Tao, Dacheng},
  booktitle={Proceedings of the AAAI Conference on Artificial Intelligence},
  volume={39},
  number={8},
  pages={7907--7915},
  year={2025}
}

@article{shao2024visual,
  title={Visual cot: Advancing multi-modal language models with a comprehensive dataset and benchmark for chain-of-thought reasoning},
  author={Shao, Hao and Qian, Shengju and Xiao, Han and Song, Guanglu and Zong, Zhuofan and Wang, Letian and Liu, Yu and Li, Hongsheng},
  journal={Advances in Neural Information Processing Systems},
  volume={37},
  pages={8612--8642},
  year={2024}
}

@article{yu2024visrag,
  title={Visrag: Vision-based retrieval-augmented generation on multi-modality documents},
  author={Yu, Shi and Tang, Chaoyue and Xu, Bokai and Cui, Junbo and Ran, Junhao and Yan, Yukun and Liu, Zhenghao and Wang, Shuo and Han, Xu and Liu, Zhiyuan and others},
  journal={arXiv preprint arXiv:2410.10594},
  year={2024}
}

@inproceedings{fu2025llmdet,
  title={Llmdet: Learning strong open-vocabulary object detectors under the supervision of large language models},
  author={Fu, Shenghao and Yang, Qize and Mo, Qijie and Yan, Junkai and Wei, Xihan and Meng, Jingke and Xie, Xiaohua and Zheng, Wei-Shi},
  booktitle={Proceedings of the Computer Vision and Pattern Recognition Conference},
  pages={14987--14997},
  year={2025}
}

@article{jin2024long,
  title={Long-context llms meet rag: Overcoming challenges for long inputs in rag},
  author={Jin, Bowen and Yoon, Jinsung and Han, Jiawei and Arik, Sercan O},
  journal={arXiv preprint arXiv:2410.05983},
  year={2024}
}

@inproceedings{zhang2024mm,
  title={MM-LLMs: Recent Advances in MultiModal Large Language Models},
  author={Zhang, Duzhen and Yu, Yahan and Dong, Jiahua and Li, Chenxing and Su, Dan and Chu, Chenhui and Yu, Dong},
  booktitle={Findings of the Association for Computational Linguistics ACL 2024},
  pages={12401--12430},
  year={2024}
}

@article{lu2024deepseek,
  title={Deepseek-vl: towards real-world vision-language understanding},
  author={Lu, Haoyu and Liu, Wen and Zhang, Bo and Wang, Bingxuan and Dong, Kai and Liu, Bo and Sun, Jingxiang and Ren, Tongzheng and Li, Zhuoshu and Yang, Hao and others},
  journal={arXiv preprint arXiv:2403.05525},
  year={2024}
}

@article{zhu2025internvl3,
  title={Internvl3: Exploring advanced training and test-time recipes for open-source multimodal models},
  author={Zhu, Jinguo and Wang, Weiyun and Chen, Zhe and Liu, Zhaoyang and Ye, Shenglong and Gu, Lixin and Tian, Hao and Duan, Yuchen and Su, Weijie and Shao, Jie and others},
  journal={arXiv preprint arXiv:2504.10479},
  year={2025}
}

@inproceedings{chen2024internvl,
  title={Internvl: Scaling up vision foundation models and aligning for generic visual-linguistic tasks},
  author={Chen, Zhe and Wu, Jiannan and Wang, Wenhai and Su, Weijie and Chen, Guo and Xing, Sen and Zhong, Muyan and Zhang, Qinglong and Zhu, Xizhou and Lu, Lewei and others},
  booktitle={Proceedings of the IEEE/CVF conference on computer vision and pattern recognition},
  pages={24185--24198},
  year={2024}
}

@article{li2025look,
  title={Look Less, Reason More: Rollout-Guided Adaptive Pixel-Space Reasoning},
  author={Li, Xuchen and Li, Xuzhao and Gao, Jiahui and Pi, Renjie and Hu, Shiyu and Zhang, Wentao},
  journal={arXiv preprint arXiv:2510.01681},
  year={2025}
}

@article{bai2025qwen3,
  title={Qwen3-vl technical report},
  author={Bai, Shuai and Cai, Yuxuan and Chen, Ruizhe and Chen, Keqin and Chen, Xionghui and Cheng, Zesen and Deng, Lianghao and Ding, Wei and Gao, Chang and Ge, Chunjiang and others},
  journal={arXiv preprint arXiv:2511.21631},
  year={2025}
}

@article{hurst2024gpt,
  title={Gpt-4o system card},
  author={Hurst, Aaron and Lerer, Adam and Goucher, Adam P and Perelman, Adam and Ramesh, Aditya and Clark, Aidan and Ostrow, AJ and Welihinda, Akila and Hayes, Alan and Radford, Alec and others},
  journal={arXiv preprint arXiv:2410.21276},
  year={2024}
}

@article{team2024gemini,
  title={Gemini 1.5: Unlocking multimodal understanding across millions of tokens of context},
  author={Team, Gemini and Georgiev, Petko and Lei, Ving Ian and Burnell, Ryan and Bai, Libin and Gulati, Anmol and Tanzer, Garrett and Vincent, Damien and Pan, Zhufeng and Wang, Shibo and others},
  journal={arXiv preprint arXiv:2403.05530},
  year={2024}
}

@article{team2025evaluating,
  title={Evaluating Gemini in an arena for learning},
  author={Team, LearnLM and Modi, Abhinit and Veerubhotla, Aditya Srikanth and Rysbek, Aliya and Huber, Andrea and Anand, Ankit and Bhoopchand, Avishkar and Wiltshire, Brett and Gillick, Daniel and Kasenberg, Daniel and others},
  journal={arXiv preprint arXiv:2505.24477},
  year={2025}
}

@article{fu2024mme,
  title={Mme-survey: A comprehensive survey on evaluation of multimodal llms},
  author={Fu, Chaoyou and Zhang, Yi-Fan and Yin, Shukang and Li, Bo and Fang, Xinyu and Zhao, Sirui and Duan, Haodong and Sun, Xing and Liu, Ziwei and Wang, Liang and others},
  journal={arXiv preprint arXiv:2411.15296},
  year={2024}
}

@article{ge2024convllava,
  title={Convllava: Hierarchical backbones as visual encoder for large multimodal models},
  author={Ge, Chunjiang and Cheng, Sijie and Wang, Ziming and Yuan, Jiale and Gao, Yuan and Song, Jun and Song, Shiji and Huang, Gao and Zheng, Bo},
  journal={arXiv preprint arXiv:2405.15738},
  year={2024}
}

@article{yue2025does,
  title={Does reinforcement learning really incentivize reasoning capacity in llms beyond the base model?},
  author={Yue, Yang and Chen, Zhiqi and Lu, Rui and Zhao, Andrew and Wang, Zhaokai and Song, Shiji and Huang, Gao},
  journal={arXiv preprint arXiv:2504.13837},
  year={2025}
}

@article{yu2025zoom,
  title={Zoom-Refine: Boosting High-Resolution Multimodal Understanding via Localized Zoom and Self-Refinement},
  author={Yu, Xuan and Guan, Dayan and Gu, Yanfeng},
  journal={arXiv preprint arXiv:2506.01663},
  year={2025}
}

@article{zhong2025focus,
  title={FOCUS: Internal MLLM representations for efficient fine-grained visual question answering},
  author={Zhong, Liangyu and Rosenthal, Fabio and Sicking, Joachim and H{\"u}ger, Fabian and Bagdonat, Thorsten and Gottschalk, Hanno and Schwinn, Leo},
  journal={arXiv preprint arXiv:2506.21710},
  year={2025}
}

@article{liu2025hide,
  title={HiDe: Rethinking The Zoom-IN method in High Resolution MLLMs via Hierarchical Decoupling},
  author={Liu, Xianjie and Hu, Yiman and Zou, Yixiong and Wu, Liang and Xu, Jian and Zheng, Bo},
  journal={arXiv preprint arXiv:2510.00054},
  year={2025}
}

@article{luan2024textcot,
  title={TextCoT: Zoom-In for Enhanced Multimodal Text-Rich Image Understanding},
  author={Luan, Bozhi and Feng, Hao and Chen, Hong and Wang, Yonghui and Zhou, Wengang and Li, Houqiang},
  journal={ACM Transactions on Multimedia Computing, Communications and Applications},
  year={2024},
  publisher={ACM New York, NY}
}

@article{chu2025sft,
  title={Sft memorizes, rl generalizes: A comparative study of foundation model post-training},
  author={Chu, Tianzhe and Zhai, Yuexiang and Yang, Jihan and Tong, Shengbang and Xie, Saining and Schuurmans, Dale and Le, Quoc V and Levine, Sergey and Ma, Yi},
  journal={arXiv preprint arXiv:2501.17161},
  year={2025}
}

@article{li2024llava,
  title={Llava-onevision: Easy visual task transfer},
  author={Li, Bo and Zhang, Yuanhan and Guo, Dong and Zhang, Renrui and Li, Feng and Zhang, Hao and Zhang, Kaichen and Zhang, Peiyuan and Li, Yanwei and Liu, Ziwei and others},
  journal={arXiv preprint arXiv:2408.03326},
  year={2024}
}

@article{dong2024internlm,
  title={Internlm-xcomposer2-4khd: A pioneering large vision-language model handling resolutions from 336 pixels to 4k hd},
  author={Dong, Xiaoyi and Zhang, Pan and Zang, Yuhang and Cao, Yuhang and Wang, Bin and Ouyang, Linke and Zhang, Songyang and Duan, Haodong and Zhang, Wenwei and Li, Yining and others},
  journal={Advances in Neural Information Processing Systems},
  volume={37},
  pages={42566--42592},
  year={2024}
}

@article{zhang2024beyond,
  title={Beyond llava-hd: Diving into high-resolution large multimodal models},
  author={Zhang, Yi-Fan and Wen, Qingsong and Fu, Chaoyou and Wang, Xue and Zhang, Zhang and Wang, Liang and Jin, Rong},
  journal={arXiv preprint arXiv:2406.08487},
  year={2024}
}

@article{luo2024feast,
  title={Feast your eyes: Mixture-of-resolution adaptation for multimodal large language models},
  author={Luo, Gen and Zhou, Yiyi and Zhang, Yuxin and Zheng, Xiawu and Sun, Xiaoshuai and Ji, Rongrong},
  journal={arXiv preprint arXiv:2403.03003},
  year={2024}
}

@inproceedings{DBLP:conf/iclr/LuoZ0ZSJ25,
  author       = {Gen Luo and
                  Yiyi Zhou and
                  Yuxin Zhang and
                  Xiawu Zheng and
                  Xiaoshuai Sun and
                  Rongrong Ji},
  title        = {Feast Your Eyes: Mixture-of-Resolution Adaptation for Multimodal Large
                  Language Models},
  booktitle    = {The Thirteenth International Conference on Learning Representations,
                  {ICLR} 2025, Singapore, April 24-28, 2025},
  publisher    = {OpenReview.net},
  year         = {2025},
  url          = {https://openreview.net/forum?id=1EnpStvBU8},
  bibsource    = {dblp computer science bibliography, https://dblp.org}
}

@article{DBLP:journals/chinaf/ChenWTYGCTHLMMWDYGHSJXW24,
  author       = {Zhe Chen and
                  Weiyun Wang and
                  Hao Tian and
                  Shenglong Ye and
                  Zhangwei Gao and
                  Erfei Cui and
                  Wenwen Tong and
                  Kongzhi Hu and
                  Jiapeng Luo and
                  Zheng Ma and
                  Ji Ma and
                  Jiaqi Wang and
                  Xiaoyi Dong and
                  Hang Yan and
                  Hewei Guo and
                  Conghui He and
                  Botian Shi and
                  Zhenjiang Jin and
                  Chao Xu and
                  Bin Wang and
                  Xingjian Wei and
                  Wei Li and
                  Wenjian Zhang and
                  Bo Zhang and
                  Pinlong Cai and
                  Licheng Wen and
                  Xiangchao Yan and
                  Min Dou and
                  Lewei Lu and
                  Xizhou Zhu and
                  Tong Lu and
                  Dahua Lin and
                  Yu Qiao and
                  Jifeng Dai and
                  Wenhai Wang},
  title        = {How far are we to GPT-4V? Closing the gap to commercial multimodal
                  models with open-source suites},
  journal      = {Sci. China Inf. Sci.},
  volume       = {67},
  number       = {12},
  year         = {2024},
  url          = {https://doi.org/10.1007/s11432-024-4231-5},
  doi          = {10.1007/S11432-024-4231-5},
  bibsource    = {dblp computer science bibliography, https://dblp.org}
}

@article{DBLP:journals/corr/abs-2403-04652,
  author       = {Alex Young and
                  Bei Chen and
                  Chao Li and
                  Chengen Huang and
                  Ge Zhang and
                  Guanwei Zhang and
                  Heng Li and
                  Jiangcheng Zhu and
                  Jianqun Chen and
                  Jing Chang and
                  Kaidong Yu and
                  Peng Liu and
                  Qiang Liu and
                  Shawn Yue and
                  Senbin Yang and
                  Shiming Yang and
                  Tao Yu and
                  Wen Xie and
                  Wenhao Huang and
                  Xiaohui Hu and
                  Xiaoyi Ren and
                  Xinyao Niu and
                  Pengcheng Nie and
                  Yuchi Xu and
                  Yudong Liu and
                  Yue Wang and
                  Yuxuan Cai and
                  Zhenyu Gu and
                  Zhiyuan Liu and
                  Zonghong Dai},
  title        = {Yi: Open Foundation Models by 01.AI},
  journal      = {CoRR},
  volume       = {abs/2403.04652},
  year         = {2024},
  url          = {https://doi.org/10.48550/arXiv.2403.04652},
  doi          = {10.48550/ARXIV.2403.04652},
  eprinttype    = {arXiv},
  eprint       = {2403.04652},
  bibsource    = {dblp computer science bibliography, https://dblp.org}
}

@misc{bai2023qwenvlversatilevisionlanguagemodel,
      title={Qwen-VL: A Versatile Vision-Language Model for Understanding, Localization, Text Reading, and Beyond}, 
      author={Jinze Bai and Shuai Bai and Shusheng Yang and Shijie Wang and Sinan Tan and Peng Wang and Junyang Lin and Chang Zhou and Jingren Zhou},
      year={2023},
      eprint={2308.12966},
      archivePrefix={arXiv},
      primaryClass={cs.CV},
      url={https://arxiv.org/abs/2308.12966}, 
}

@article{zhang2024mme,
  title={Mme-realworld: Could your multimodal llm challenge high-resolution real-world scenarios that are difficult for humans?},
  author={Zhang, Yi-Fan and Zhang, Huanyu and Tian, Haochen and Fu, Chaoyou and Zhang, Shuangqing and Wu, Junfei and Li, Feng and Wang, Kun and Wen, Qingsong and Zhang, Zhang and others},
  journal={arXiv preprint arXiv:2408.13257},
  year={2024}
}

@article{zhang2025thyme,
  title={Thyme: Think beyond images},
  author={Zhang, Yi-Fan and Lu, Xingyu and Yin, Shukang and Fu, Chaoyou and Chen, Wei and Hu, Xiao and Wen, Bin and Jiang, Kaiyu and Liu, Changyi and Zhang, Tianke and others},
  journal={arXiv preprint arXiv:2508.11630},
  year={2025}
}

@article{wang2025pixel,
  title={Pixel reasoner: Incentivizing pixel-space reasoning with curiosity-driven reinforcement learning},
  author={Wang, Haozhe and Su, Alex and Ren, Weiming and Lin, Fangzhen and Chen, Wenhu},
  journal={arXiv preprint arXiv:2505.15966},
  year={2025}
}

@article{li2025towards,
  title={Towards a holistic framework for multimodal LLM in 3D brain CT radiology report generation},
  author={Li, Cheng-Yi and Chang, Kao-Jung and Yang, Cheng-Fu and Wu, Hsin-Yu and Chen, Wenting and Bansal, Hritik and Chen, Ling and Yang, Yi-Ping and Chen, Yu-Chun and Chen, Shih-Pin and others},
  journal={Nature Communications},
  volume={16},
  number={1},
  pages={2258},
  year={2025},
  publisher={Nature Publishing Group UK London}
}

@article{zambrano2025clinically,
  title={A clinically accessible small multimodal radiology model and evaluation metric for chest X-ray findings},
  author={Zambrano Chaves, Juan Manuel and Huang, Shih-Cheng and Xu, Yanbo and Xu, Hanwen and Usuyama, Naoto and Zhang, Sheng and Wang, Fei and Xie, Yujia and Khademi, Mahmoud and Yang, Ziyi and others},
  journal={Nature Communications},
  volume={16},
  number={1},
  pages={3108},
  year={2025},
  publisher={Nature Publishing Group UK London}
}

@inproceedings{wang2024videoagent,
  title={Videoagent: Long-form video understanding with large language model as agent},
  author={Wang, Xiaohan and Zhang, Yuhui and Zohar, Orr and Yeung-Levy, Serena},
  booktitle={European Conference on Computer Vision},
  pages={58--76},
  year={2024},
  organization={Springer}
}

@inproceedings{wang2025videotree,
  title={Videotree: Adaptive tree-based video representation for llm reasoning on long videos},
  author={Wang, Ziyang and Yu, Shoubin and Stengel-Eskin, Elias and Yoon, Jaehong and Cheng, Feng and Bertasius, Gedas and Bansal, Mohit},
  booktitle={Proceedings of the Computer Vision and Pattern Recognition Conference},
  pages={3272--3283},
  year={2025}
}

@article{hwangemma,
  title={EMMA: End-to-End Multimodal Model for Autonomous Driving},
  author={Hwang, Jyh-Jing and Xu, Runsheng and Lin, Hubert and Hung, Wei-Chih and Ji, Jingwei and Choi, Kristy and Huang, Di and He, Tong and Covington, Paul and Sapp, Benjamin and others},
  journal={Transactions on Machine Learning Research}
}

@inproceedings{xing2025openemma,
  title={Openemma: Open-source multimodal model for end-to-end autonomous driving},
  author={Xing, Shuo and Qian, Chengyuan and Wang, Yuping and Hua, Hongyuan and Tian, Kexin and Zhou, Yang and Tu, Zhengzhong},
  booktitle={Proceedings of the Winter Conference on Applications of Computer Vision},
  pages={1001--1009},
  year={2025}
}

@article{nguyen2024yo,
  title={Yo'llava: Your personalized language and vision assistant},
  author={Nguyen, Thao and Liu, Haotian and Li, Yuheng and Cai, Mu and Ojha, Utkarsh and Lee, Yong Jae},
  journal={Advances in Neural Information Processing Systems},
  volume={37},
  pages={40913--40951},
  year={2024}
}

@article{an2025unictokens,
  title={Unictokens: Boosting personalized understanding and generation via unified concept tokens},
  author={An, Ruichuan and Yang, Sihan and Zhang, Renrui and Shen, Zijun and Lu, Ming and Dai, Gaole and Liang, Hao and Guo, Ziyu and Yan, Shilin and Luo, Yulin and others},
  journal={arXiv preprint arXiv:2505.14671},
  year={2025}
}

@article{liu2024robomamba,
  title={Robomamba: Efficient vision-language-action model for robotic reasoning and manipulation},
  author={Liu, Jiaming and Liu, Mengzhen and Wang, Zhenyu and An, Pengju and Li, Xiaoqi and Zhou, Kaichen and Yang, Senqiao and Zhang, Renrui and Guo, Yandong and Zhang, Shanghang},
  journal={Advances in Neural Information Processing Systems},
  volume={37},
  pages={40085--40110},
  year={2024}
}

@inproceedings{li2024manipllm,
  title={Manipllm: Embodied multimodal large language model for object-centric robotic manipulation},
  author={Li, Xiaoqi and Zhang, Mingxu and Geng, Yiran and Geng, Haoran and Long, Yuxing and Shen, Yan and Zhang, Renrui and Liu, Jiaming and Dong, Hao},
  booktitle={Proceedings of the IEEE/CVF Conference on Computer Vision and Pattern Recognition},
  pages={18061--18070},
  year={2024}
}

@article{li2025mini,
  title={Mini-gemini: Mining the potential of multi-modality vision language models},
  author={Li, Yanwei and Zhang, Yuechen and Wang, Chengyao and Zhong, Zhisheng and Chen, Yixin and Chu, Ruihang and Liu, Shaoteng and Jia, Jiaya},
  journal={IEEE Transactions on Pattern Analysis and Machine Intelligence},
  year={2025},
  publisher={IEEE}
}
}

\renewcommand{\thesection}{\Alph{section}}

\maketitlesupplementary
\setcounter{section}{0}
\section{Implement Details of \textit{MRD}}
\setcounter{page}{1}

\label{sec:Implementation Details}

Given an input high-resolution image \textit{I}, we first partition it into fixed-size local crops, with crop dimensions predefined according to the native resolution of the input image, following the protocol of \textit{RAP}~\cite{wang2025retrieval}. Specifically for our proposed \textbf{\textit{MRD}} framework, we set the crop resolution to 112, 224, and 448 for the $V^*$ \textit{Bench}, \textit{HR-Bench-4K}, and \textit{HR-Bench-8K} benchmarks, respectively, to match the native resolution of each dataset. For the multi-resolution semantic fusion module in our framework, we fix the resolution ratio between the high-resolution and low-resolution branches to $k=2$ for all experiments.

For the sliding window detection pipeline, we configure the window size and stride to strike a balance between inference efficiency and detection accuracy: we set the (window size, stride) pairs to (1232, 896) for $V^*$ \textit{Bench}, (2240, 1792) for \textit{HR-Bench-4K}, and (3136, 2688) for \textit{HR-Bench-8K}, respectively. We set the default detection confidence threshold to 0.3 to filter out low-quality bounding box predictions, and fix the weight $w$ of the detection confidence map to 0.4 for the semantic detection map fusion step. For the subsequent \textit{Retrieval-Exploration Search} pipeline, we adopt the same hyperparameter settings as the \textit{RAP} baseline method, with one critical exception: our method achieves peak performance with fewer search iterations, as validated in Sec.~\ref{sec:Maximum Search Steps}. Accordingly, we fix the maximum number of search steps to 50 across all our experiments.

For all subsequent hyperparameter ablation studies, all other components of our proposed \textbf{\textit{MRD}} framework are fixed to the above default settings unless explicitly specified otherwise.

\section{Additional Ablation Studies}
\label{sec:ablation_studies}

To further understand the behavior of our framework and evaluate the robustness of the proposed \textbf{\textit{MRD}} module, we conduct additional ablation studies by varying several key hyperparameters. These experiments aim to analyze how different parameter settings influence the overall performance of the system.

Specifically, we vary important hyperparameters such as \textit{crop resolution}, \textit{more resolution fusion}, \textit{maximum search steps}, \textit{detection weight}, sliding window size, and detection confidence threshold (Sec.~\ref{subsec:ablation_crop_resolution} to Sec.~\ref{subsec:ablation_crop_resolution}). For each experiment, we modify only one hyperparameter while keeping all others fixed, allowing us to isolate the effect of each factor on model performance.

All experiments are conducted on the $V^*$ \textit{Bench}  using two representative multimodal large language models, LLaVA-ov-0.5B and LLaVA-v1.5-7B. This setup enables us to evaluate the sensitivity of the proposed method across different model scales. Unless otherwise specified, all remaining hyperparameters follow the default settings described in Sec.~\ref{sec:Implementation Details}. The detailed results are presented in the following subsections.

% \section{Additional Ablation Studies}
% \label{sec:ablation_studies}
% To systematically analyze how key hyperparameters affect the performance of our framework, we conduct comprehensive ablation studies on the proposed \textbf{\textit{MRD}} by varying critical hyperparameters, including crop resolution, maximum search steps, multi-reslution fusion, detection confidence map weight, sliding window size, and detection confidence threshold. All experiments are conducted on the $V^*$ \textit{Bench} using both the LLaVA-ov-0.5B and LLaVA-v1.5-7B models. Unless explicitly stated otherwise, all other hyperparameters are fixed to the default settings detailed in Sec.~\ref{sec:Implementation Details}.
\subsection{Effect of Crop Resolution}
\label{subsec:ablation_crop_resolution}

\begin{figure*}[t]

  \centering
   \includegraphics[width=0.9\linewidth]{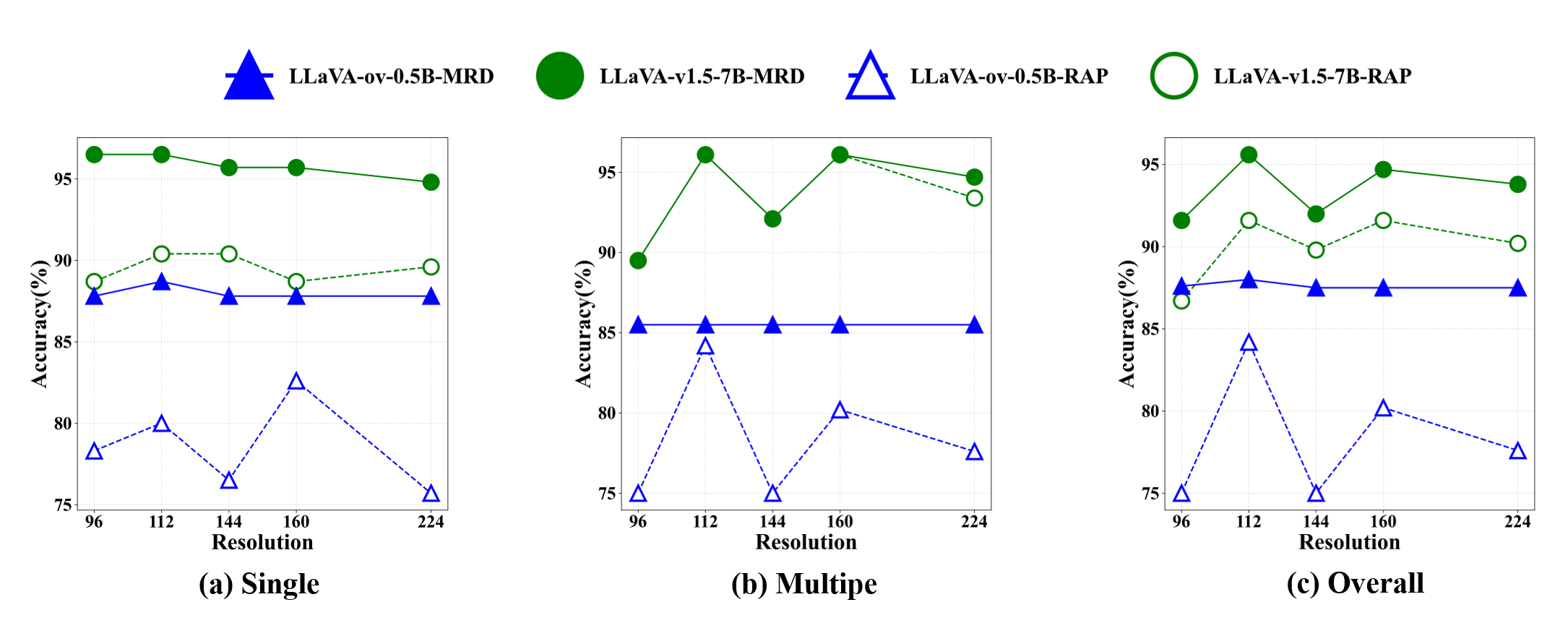}

   \caption{ The effect of the resolution of image crops on MLLMs performance. Single and Multiple represent the attribute recognition and spatial reasoning tasks in $V^{*}$ Bench. (a) Single-object Task. (b) Multi-object Task. (c) Overall Performance.}. 
   \label{fig:Resolution_effect}
   
\end{figure*}

To systematically investigate the impact of crop resolution on performance, we conduct ablation experiments under a range of crop size settings. The results are shown in Fig.~\ref{fig:Resolution_effect}.
For the single-object task (Fig.~\ref{fig:Resolution_effect} (a)), the proposed \textbf{\textit{MRD}} demonstrates consistently stable performance across all tested resolutions for both backbones, with only minor fluctuations. In contrast, the \textit{RAP} baseline exhibits noticeable performance instability under varying crop resolutions, particularly when using the lightweight LLaVA-ov-0.5B model. For the multi-object task (Fig.~\ref{fig:Resolution_effect} (b)), the performance gap between \textbf{\textit{MRD}} and \textit{RAP} becomes smaller when using the stronger LLaVA-v1.5-7B backbone. However, when switching to the smaller LLaVA-ov-0.5B model, \textbf{\textit{MRD}} still maintains stable performance across different resolutions, further demonstrating its robustness to resolution changes. Overall, \textbf{\textit{MRD}} consistently outperforms the \textit{RAP} baseline across all tested crop resolutions and backbone configurations.

These results suggest our \textbf{\textit{MRD}} effectively mitigate the object fragmentation issue that often arises when objects are split across adjacent crops under different resolutions. As a result, our framework shows reduced sensitivity to crop resolution and achieves more stable performance, particularly in the single-object setting.

\subsection{Effect of More Resolution Fusion}
\label{subsec:More_Resolution_Fusion}

To further investigate the impact of using additional resolution scales in the \textit{Multi-resolution Semantic Fusion} module, we conduct experiments by incorporating multi-scale fusion into the RAP framework. Specifically, we evaluate three scale settings with $k=1,2,4$, while keeping the crop resolution fixed at 112 for all image patches. Experiments are conducted on the $V^*$ \textit{Bench} using two representative backbones, LLaVA-ov-0.5B and LLaVA-v1.5-7B. The results are summarized in Tab.~\ref{tab:multi_res_effect} For the  LLaVA-ov-0.5B model, fusing three resolution scales ($k=1,2,4$) achieves better performance than using two scales or a single scale for both the single-object and multi-object tasks. In contrast, for the LLaVA-v1.5-7B backbone, using two scales ($k=1,2$) leads to better performance than incorporating an additional resolution level.

These results suggest that introducing more resolution scales does not necessarily lead to consistent performance improvements across different backbone models. However, multi-resolution fusion consistently outperforms the single-resolution setting in all cases, indicating the effectiveness of leveraging complementary information from different crop scales. Considering both performance and computational efficiency, we adopt a two-resolution scheme ($k=1,2$) in practice.

\begin{table}[h]
\caption{Effect of more resolution fusion on $V*$ \textit{Bench}.}
\centering
\small
\setlength{\tabcolsep}{6pt}
\begin{tabular}{lccc}
\toprule
\multirow{2}{*}{\textbf{Method}} & \multicolumn{3}{c}{\textit{\textbf{V* Bench}}} \\
\cmidrule(lr){2-4}
 & \textbf{Attribute} & \textbf{Spatial} & \textbf{Overall} \\
\midrule

RAP (LLaVA-ov-0.5B) & 80.0 & 84.2 & 83.6 \\
+Multi-Res (k=1,2) & \textbf{83.5} & 85.5 & 85.8 \\
+Multi-Res (k=1,4) & 80.0 & 86.8 & 84.4 \\
+Multi-Res (k=1,2,4) & \textbf{83.5} & \textbf{88.2} & \textbf{86.7} \\

\midrule

RAP (LLaVA-v1.5-7B) & 90.4 & \textbf{96.1} & 91.1 \\
+Multi-Res (k=1,2) & \textbf{94.8} & \textbf{96.1} & \textbf{94.2} \\
+Multi-Res (k=1,4) & 93.9 & 94.7 & 93.3 \\
+Multi-Res (k=1,2,4) & 93.9 & \textbf{96.1} & \textbf{94.2} \\

\bottomrule
\end{tabular}
\label{tab:multi_res_effect}
\end{table}

\subsection{Effect of Maximum Search Steps}
\label{sec:Maximum Search Steps}

\begin{figure*}[t]
  \centering
   \includegraphics[width=0.9\linewidth]{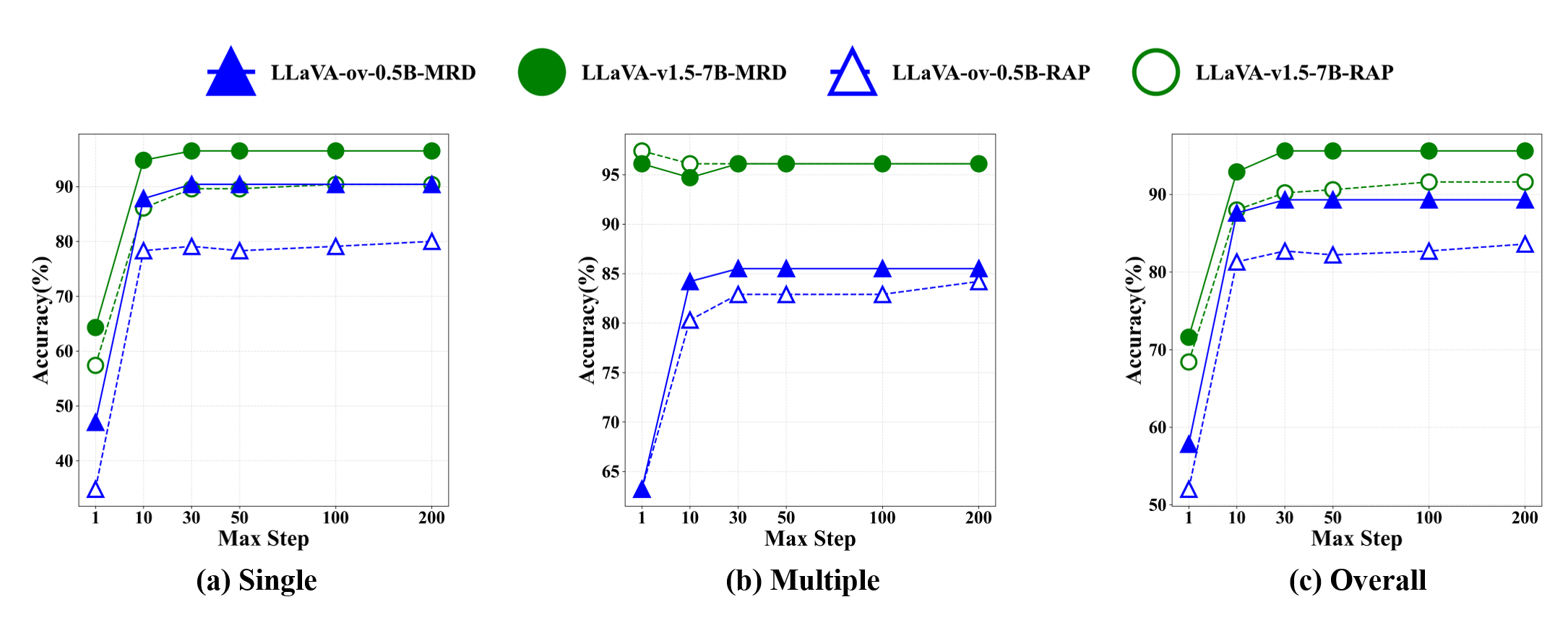}

   \caption{ The effect of the maximum search steps in \textit{\textbf{MRD}} and \textit{RAP}.}
   \label{fig:Max step effect}
\end{figure*}

The performance of \textbf{\textit{MRD}} and \textit{RAP} under different maximum search steps is illustrated in Fig.~\ref{fig:Max step effect}.For the single-object task (Fig.~\ref{fig:Max step effect} (a)), \textbf{\textit{MRD}} consistently outperforms \textit{RAP} across all tested maximum step settings on both the LLaVA-ov-0.5B and LLaVA-v1.5-7B backbones, demonstrating stable performance under varying search budgets. For the multi-object task (Fig.~\ref{fig:Max step effect} (b)), \textbf{\textit{MRD}} is slightly inferior to \textit{RAP} only when using the LLaVA-v1.5-7B backbone with a very small number of maximum steps. As the maximum step increases, \textbf{\textit{MRD}} quickly surpasses \textit{RAP} and maintains superior performance. Overall, \textbf{\textit{MRD}} achieves better results than \textit{RAP} across most settings. Notably, \textbf{\textit{MRD}} with the lightweight LLaVA-ov-0.5B backbone achieves performance that is only marginally lower than \textit{RAP} with the much larger LLaVA-v1.5-7B model.

Another important observation is that \textbf{\textit{MRD}} reaches its peak performance with a relatively small number of maximum search steps (Max Step = 50). This indicates that our method can achieve strong performance with a limited search budget, reducing search time while maintaining high accuracy in practical scenarios.

\subsection{Effect of Detection Weight}

\begin{figure*}[t]
  \centering
   \includegraphics[width=0.9\linewidth]{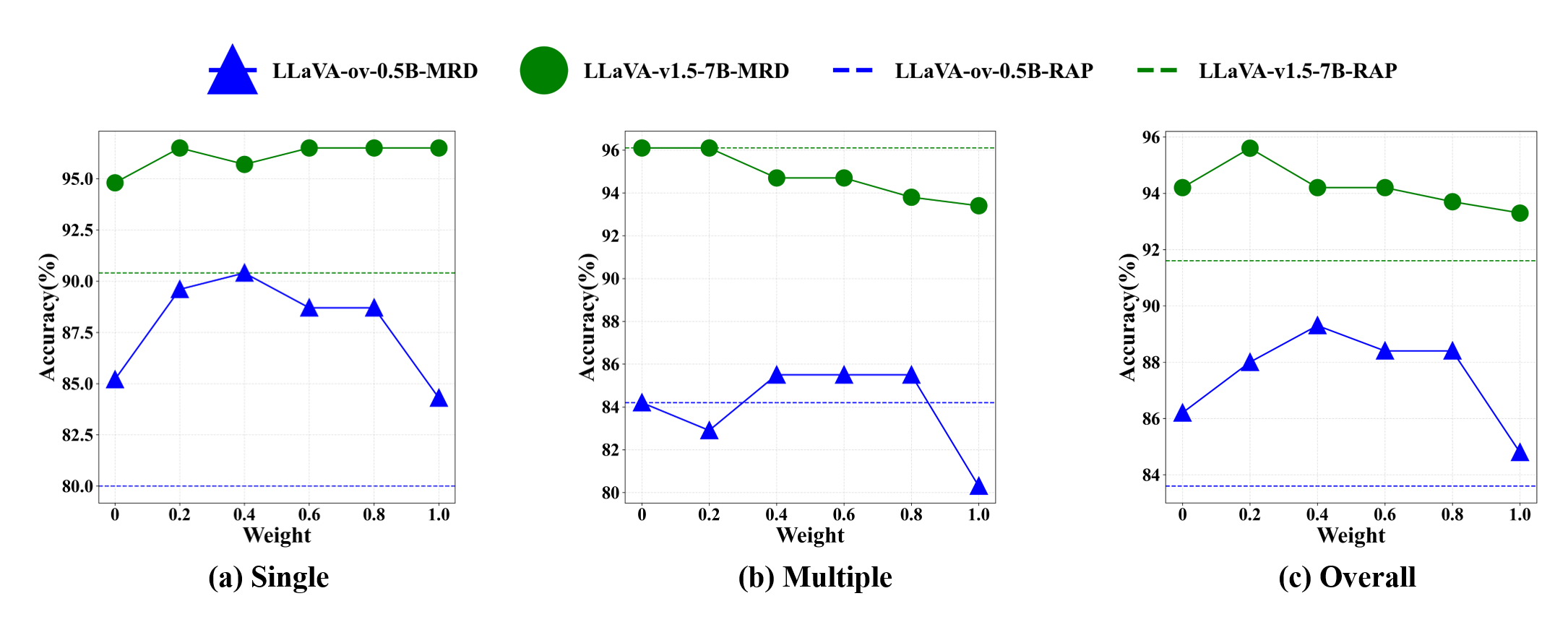}

   \caption{ The effect of the detection weight in \textit{\textbf{MRD}}.}
   \label{fig:detection weight effect}
\end{figure*}

The results under different detection weight settings are presented in Fig.~\ref{fig:detection weight effect}. We observe that relying solely on the multi-resolution semantic similarity map (weight $w$= 0) or solely on the detection map (weight $w$ = 1) does not yield optimal performance for either task. In contrast, combining the two maps leads to improved results, indicating that the semantic similarity map and the detection map provide complementary information.

Overall, the optimal detection weight varies slightly across different backbones. The lightweight LLaVA-ov-0.5B model achieves its best performance at a detection weight of 0.4, while LLaVA-v1.5-7B performs best when the detection weight is set to 0.2. These observations further validate the benefit of integrating both semantic and detection cues in our framework.

% \begin{figure*}[t]
%   \centering
%    \includegraphics[width=1\linewidth]{fig/Weight.png}

%    \caption{ The effect of the detection weight in \textbf{\textit{MRD}}.}
%    \label{fig:detection weight effect}
% \end{figure*}

% \subsection{Effect of Window Size}
% As shown in \autoref{fig:window size effect}, adopting different sliding-window sizes for object detection also affects the results. Except for the Cross Instance Task with LLaVA-ov-0.5B, using a smaller sliding-window size (Window Size = 896) generally yields better performance. This is because a smaller window reduces background interference unrelated to the target object, leading to more accurate detection results.

% However, a smaller window size also means that more windows are required to scan the entire high-resolution image, resulting in increased computational complexity and longer processing time. Therefore, to balance accuracy and efficiency, we select a larger sliding-window size, Window Size = 1232, as the default setting.

\subsection{Effect of Window Size}

\begin{figure*}[t]
  \centering
   \includegraphics[width=0.9\linewidth]{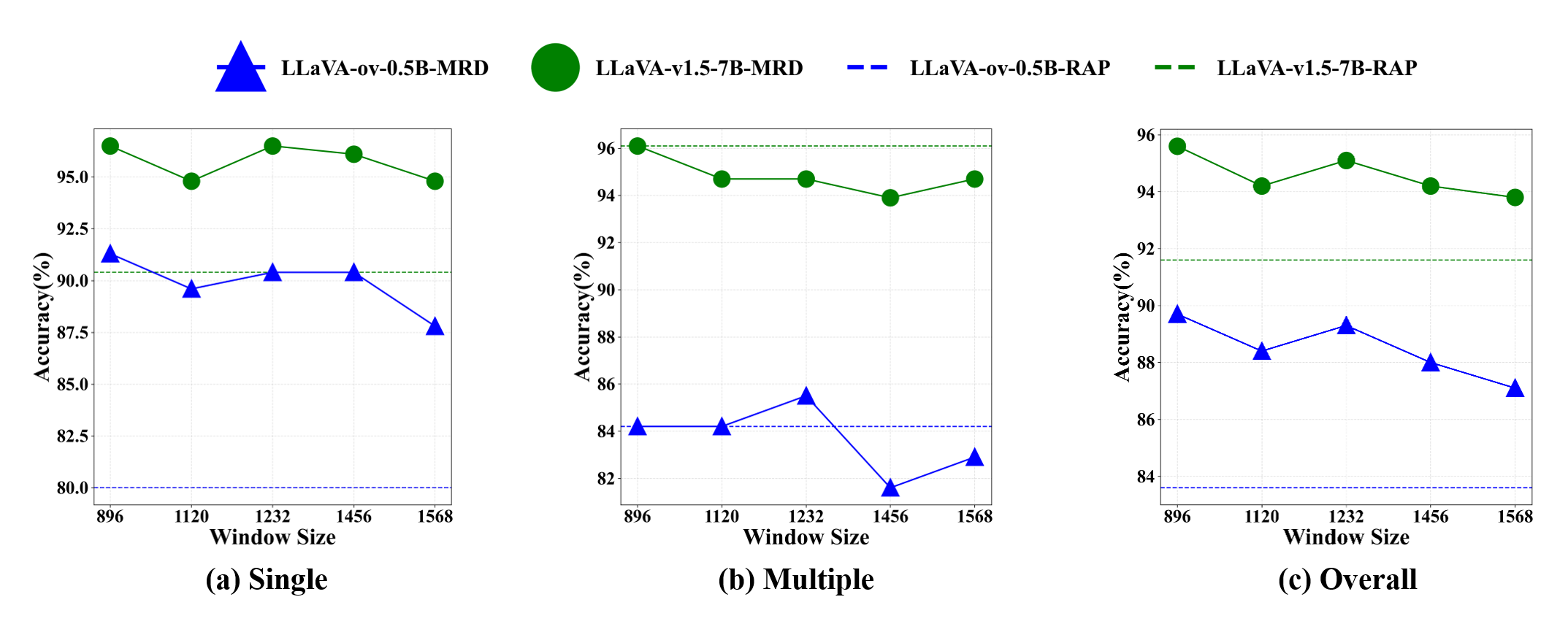}

   \caption{ The effect of the detection window size in \textbf{\textit{MRD}}.}
   \label{fig:window size effect}
\end{figure*}

As shown in Fig.~\ref{fig:window size effect}, the choice of sliding-window size for object detection also has a noticeable impact on the final performance. In most settings, using a smaller sliding-window size (Window Size = 896) leads to slightly better results, except for the multi-object task with the LLaVA-ov-0.5B backbone. This improvement can be attributed to the reduced background context within smaller windows, which helps suppress irrelevant regions and enables the detector to focus more precisely on the target objects.

However, reducing the window size also introduces additional computational costs. Specifically, smaller windows require a larger number of sliding windows to cover the entire high-resolution image, which increases both the number of detection operations and the overall processing time. As a result, although smaller windows may provide marginal accuracy gains, they lead to reduced efficiency in large-scale inference scenarios.

To balance detection accuracy and computational efficiency, we adopt a larger sliding-window size (Window Size = 1232) as the default configuration in our framework. This setting provides a favorable trade-off between performance and runtime, while still maintaining competitive detection accuracy across different tasks and backbone models.

% \subsection{Effect of Detection Confidence Threshold}

\begin{figure*}[t]
  \centering
   \includegraphics[width=0.9\linewidth]{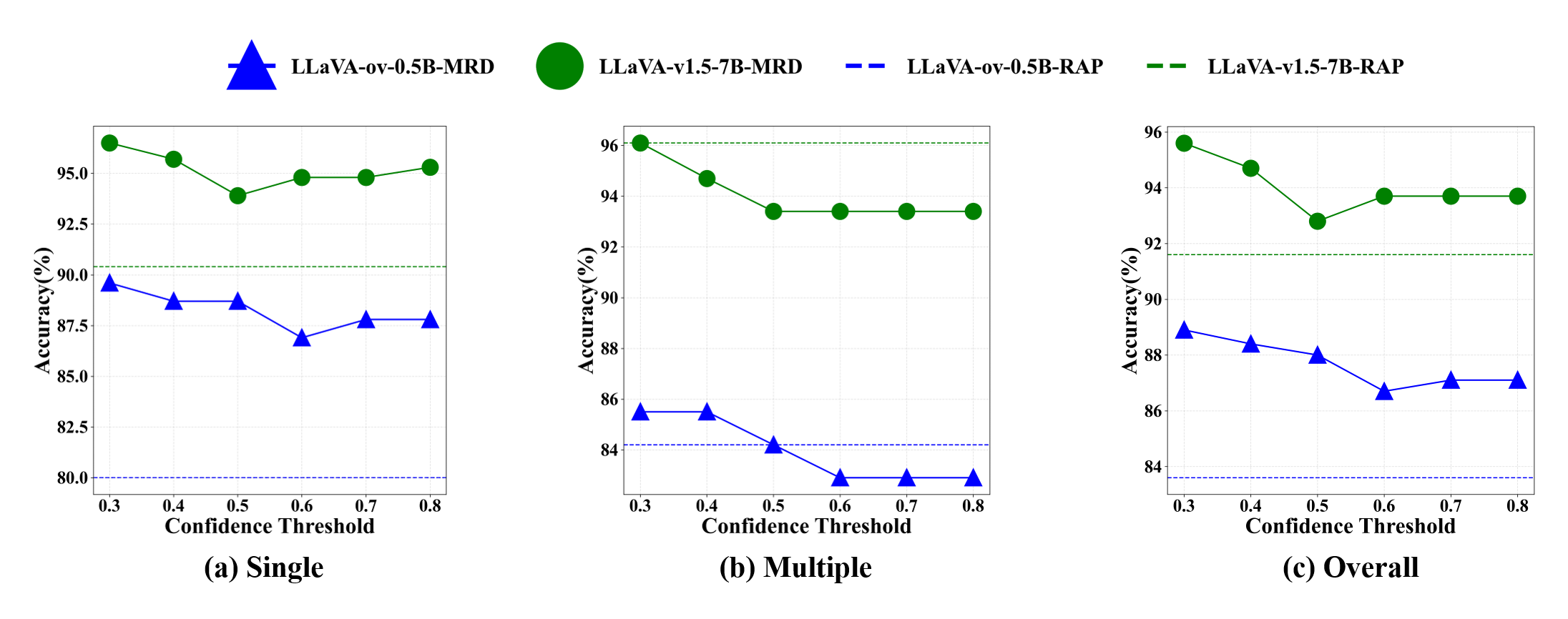}

   \caption{ The effect of the detection confidence threshold in \textbf{\textit{MRD}}.}
   \label{fig:confidence threshold effect}
\end{figure*}

\subsection{Effect of Detection Confidence Threshold}

The effect of different detection confidence thresholds is illustrated in Fig.~\ref{fig:confidence threshold effect}. We evaluate a range of threshold values to analyze how filtering detection candidates influences the overall performance.
From Fig.~\ref{fig:confidence threshold effect}, we observe that the best performance is consistently achieved when the confidence threshold is set to 0.3. This trend holds across both backbone models (LLaVA-ov-0.5B and LLaVA-v1.5-7B) and across different tasks, including the single-object task, the multi-object task, and the overall evaluation.

As the threshold increases, the performance gradually decreases in most cases. A higher threshold filters out more detection candidates, which may remove useful object proposals and reduce the coverage of potential target regions. In contrast, using a lower threshold retains more candidate regions, allowing the subsequent reasoning and fusion modules to better identify relevant objects.

Based on these observations, we adopt a confidence threshold of 0.3 as the default setting in our framework, as it consistently provides the best performance across different tasks and backbone models.

\section{More Visualization Results}
\label{sec:Case Study}

\subsection{Examples of Single-object Perception Task}

\begin{figure*}[!t]
  \centering
   \includegraphics[width=1\linewidth]{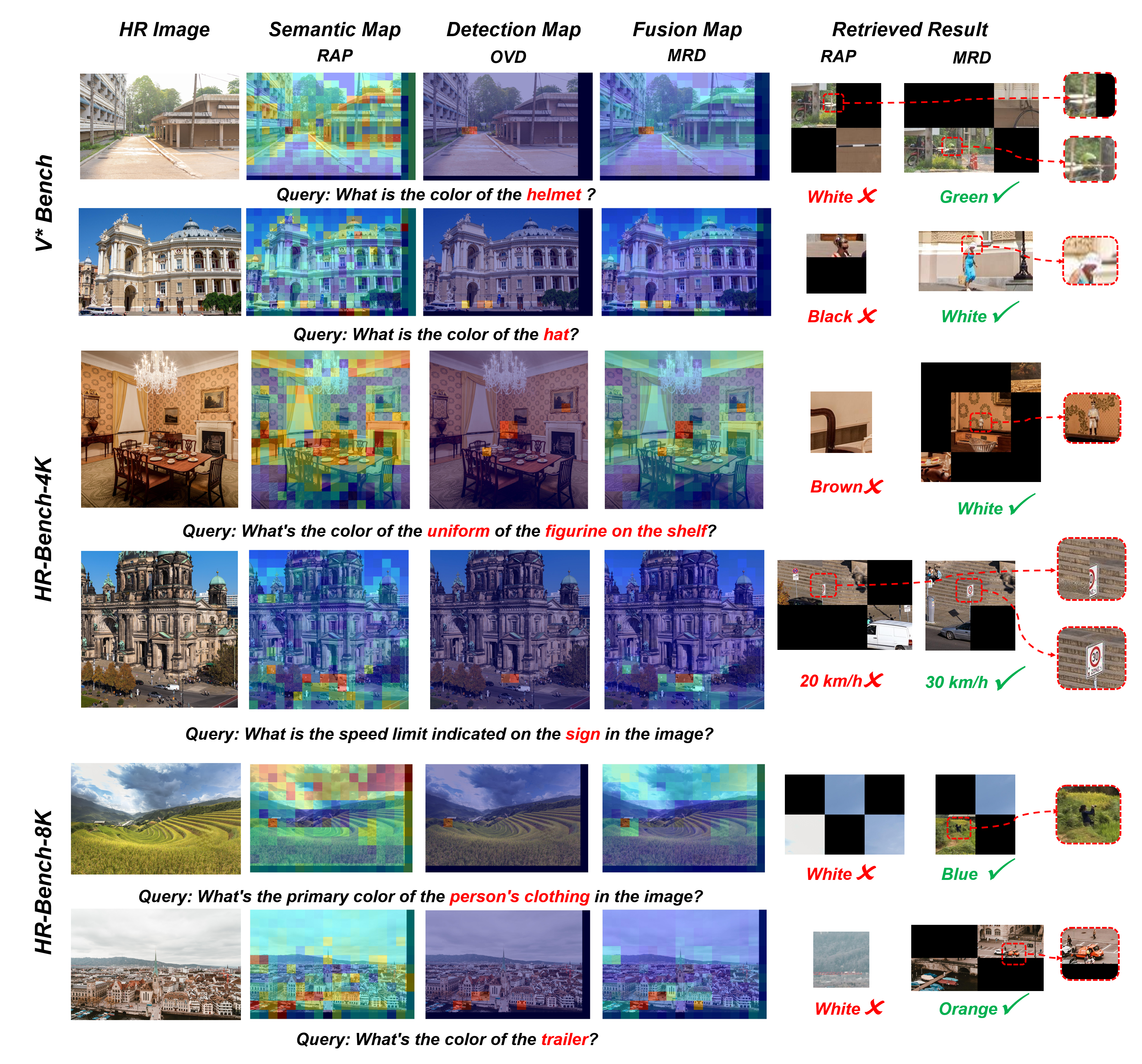}

   \caption{Qualitative examples of single-object perception task.  We conduct experiments using LLaVA-v1.5-7B on three HR Benchmarks.}
   \label{fig:Single case}
\end{figure*}

Fig.~\ref{fig:Single case} presents two qualitative examples of the single-object perception task from each HR benchmark, comparing \textbf{\textit{MRD}} with \textit{RAP} using the LLaVA-v1.5-7B backbone. From left to right, we visualize the HR image, the semantic similarity map produced by \textit{RAP}, the object detection confidence map, the semantic–detection fusion map generated by \textbf{\textit{MRD}}, and the final predictions from \textit{RAP} and \textbf{\textit{MRD}}.
From the visualization of the \textit{RAP} semantic similarity maps, it can be observed that crop-based partitioning may split a complete object across multiple image crops, resulting in inconsistent semantic similarity scores across different parts of the object. Such inconsistencies can negatively affect the subsequent retrieval process. For example, in the second case from \textit{HR-Bench4K}, \textit{RAP} retrieves only the right half of the speed-limit sign, leading to an incorrect final prediction. In addition, the similarity maps often contain false positives. For instance, in the first example from \textit{HR-Bench8K}, the sky region—although irrelevant to the query—exhibits undesirably high similarity scores.

The proposed \textbf{\textit{MRD}} addresses these issues through two key mechanisms. First, the multi-resolution semantic fusion module mitigates semantic inconsistencies caused by crop partitioning by aggregating information across multiple resolutions, which helps preserve the integrity of the target object. Second, by incorporating an object detection model to explicitly localize candidate regions, \textbf{\textit{MRD}} enhances the similarity scores of true target regions while suppressing irrelevant responses. As illustrated in Fig.~\ref{fig:Single case}, the resulting semantic–detection fusion maps exhibit clearer contrast between target objects and background regions compared with those produced by \textit{RAP}. Consequently, false positives are significantly reduced, enabling more accurate retrieval of target-related crops during the search process.

\subsection{Examples of Multi-object Perception Task}

\begin{figure*}[!t]
  \centering
   \includegraphics[width=1\linewidth]{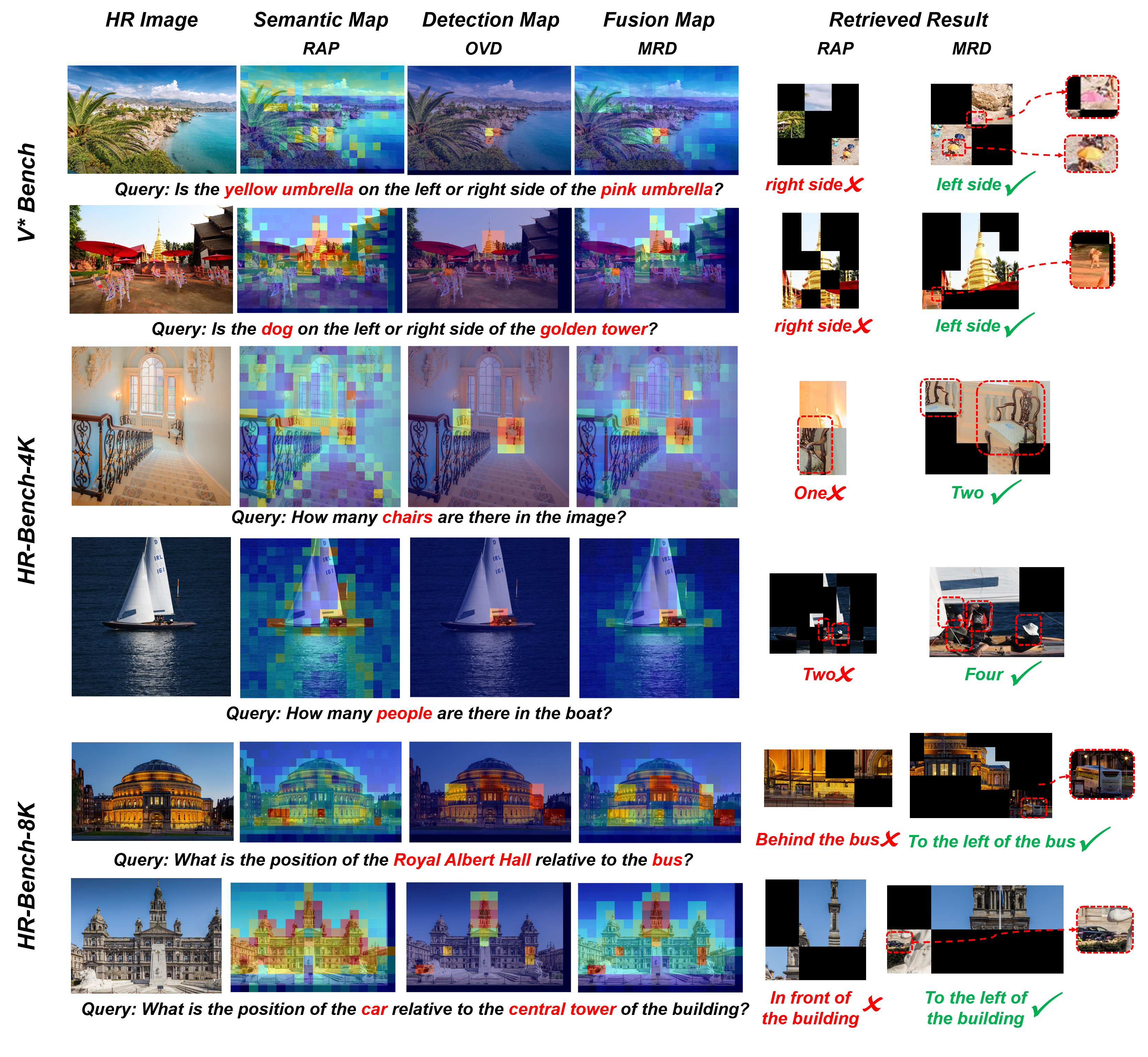}

   \caption{Qualitative examples of multi-object perception task.We conduct experiments using LLaVA-v1.5-7B on three HR Benchmarks.}
   \label{fig:Multiple case}
\end{figure*}

Fig.~\ref{fig:Multiple case} presents two qualitative examples of the multi-object perception task from each HR benchmark, comparing the performance of \textbf{\textit{MRD}} and \textit{RAP} using the LLaVA-v1.5-7B backbone. In the multi-object perception task, the retrieval results reveal that \textit{RAP} often preserves only a subset of the target objects while ignoring others when multiple objects must be localized simultaneously. For example, in the first case from $V^*$ \textit{Bench}, \textit{RAP} completely fails to retrieve the pink umbrella. This issue becomes more evident when there is a large scale discrepancy among the target objects. As shown in the second case of $V^*$ \textit{Bench} and the two examples from \textit{HRBench-8K}, \textit{RAP} tends to retain only the dominant large object while neglecting smaller objects that are also relevant to the query. Similar behavior can also be observed in counting scenarios, such as the two examples from \textit{HRBench-4K}, where \textit{RAP} identifies only a subset of the object instances, resulting in incomplete predictions.

These limitations mainly arise from the reliance on semantic similarity maps in \textit{RAP}, where smaller objects or objects with weaker semantic responses may be suppressed during the retrieval process. In contrast, \textbf{\textit{MRD}} incorporates an object detection module that explicitly localizes candidate objects before the retrieval stage. This design enables the framework to simultaneously preserve multiple target instances, including small or less prominent objects. As illustrated in Fig.~\ref{fig:Multiple case}, \textbf{\textit{MRD}} is able to retain all relevant objects more reliably, leading to more accurate predictions in cross-instance perception tasks.

\section{Limitations and Future Works}

Despite the substantial improvements achieved by \textbf{\textit{MRD}} on single-object perception tasks, the gains on multi-object scenarios are relatively more modest. We believe this limitation mainly comes from the open-vocabulary detection component, which may be less effective in complex scenes containing many small, partially occluded, or densely packed instances. In such cases, the sliding-window detection strategy may exhibit a bias toward salient or dominant objects, thereby reducing recall for less prominent targets.

Another potential limitation is that the current framework relies on fixed window settings and hand-designed fusion hyperparameters. While these settings work well in practice, they may not always be optimal across different image scales, scene complexities, or backbone models. As a result, the performance improvement on more challenging multi-object cases is still constrained by the detection quality and the coverage of candidate regions.

In future work, we plan to investigate more adaptive region exploration strategies, such as content-aware window selection and dynamic multi-scale scanning, to improve small-object recall and better handle crowded scenes. We also aim to explore stronger detection-guided fusion mechanisms that can more effectively balance semantic relevance and spatial localization. Importantly, we hope to preserve the modular and training-free nature of \textbf{\textit{MRD}}, so that it remains easy to integrate into different multimodal large language models without additional training.

\end{document}